\documentclass[review]{fcs}
\usepackage{microtype}
\usepackage{mathrsfs}
\usepackage{multirow} 
\usepackage{longtable}
\usepackage[english]{babel}
\usepackage{bm}
\usepackage{booktabs}
\usepackage{threeparttable}
\usepackage{pifont}
\usepackage{wasysym}
\setlist[itemize]{leftmargin=*}
\def\x{\bm{x}}
\def\z{\bm{z}}

\def\Z{\mathbf{Z}}
\def\L{\mathbf{L}}
\def\I{\mathbf{I}}
\def\P{\mathbf{P}}
\def\A{\mathbf{A}}
\def\D{\mathbf{D}}

\def\S{\mathbf{S}}

\def\X{\mathbf{X}}
\def\H{\mathbf{H}}

\def\i{\mathbf{i}}
\def\f{\mathbf{f}}
\def\o{\mathbf{o}}
\def\c{\mathbf{c}}
\def\u{\mathbf{u}}
\def\b{\mathbf{b}}
\def\C{\mathbf{C}}
\def\W{\mathbf{W}}

\def\z{\mathbf{z}}
\def\h{\mathbf{h}}
\def\B{\mathbf{B}}

\def\Dn{\mathbf{D}^{-\frac{1}{2}}}

\def\revision{}
\def\polish{}

\newcommand{\eat}[1]{}

\title{A survey of dynamic graph neural networks}
\author{Yanping ZHENG}
\author{Lu YI}
\author*{Zhewei WEI}
\address{Gaoling School of Artificial Intelligence, Renmin University of China, Beijing 100872, China}
\fcssetup{
  doi            = {10.1007/s11704-024-3853-2},
  received       = {\polish October 30, 2023},
  accepted       = {\polish April 23, 2024},
  corr-email     = {zhewei@ruc.edu.cn.},
}
\begin{abstract}
Graph neural networks (GNNs) have emerged as a powerful tool for effectively mining and learning from graph-structured data, with applications spanning numerous domains. However, most research focuses on static graphs, neglecting the dynamic nature of real-world networks where topologies and attributes evolve over time. By integrating sequence modeling modules into traditional GNN architectures, dynamic GNNs aim to bridge this gap, capturing the inherent temporal dependencies of dynamic graphs for a more authentic depiction of complex networks. This paper provides a comprehensive review of the fundamental concepts, key techniques, and state-of-the-art dynamic GNN models. We present the mainstream dynamic GNN models in detail and categorize models based on how temporal information is incorporated. We also discuss large-scale dynamic GNNs and pre-training techniques. Although dynamic GNNs have shown superior performance, challenges remain in scalability, handling heterogeneous information, and lack of diverse graph datasets. The paper also discusses possible future directions, such as adaptive and memory-enhanced models, inductive learning, and theoretical analysis. 
\end{abstract}
\keywords{graph neural networks, dynamic graph, temporal modeling, large-scale}

\begin{document}
% \section{Required}
% Manuscripts should be in a Word or LATEX format. The following components are required for a complete regular manuscript: \textbf{Title, Author(s), Author affiliation(s), Abstract, Keywords, Nomenclature (if needed), Main text, Acknowledgements, Competing interests, References, Appendices (if needed), Figure captions, Tables}. There is no formal limit for the length of a paper, but the editors may recommend condensation when appropriate.
\section{Introduction}
\label{sec:intro}
With the remarkable advances in deep learning across various application domains, Graph Neural Networks (GNNs) have steadily emerged as a prominent solution for addressing problems involving complex graph-structured data. In the real world, crucial information is often represented in the form of graphs, such as relationships within social networks, roads and intersections in transportation systems, and protein interaction networks in bioinformatics. Compared to static graphs, dynamic graphs provide a more realistic representation of real-world systems and networks, as they can model structural and attribute information that evolves over time~\cite{survey2020representation}.

\begin{figure*}[t]
    \centering
    \includegraphics[height=60mm]{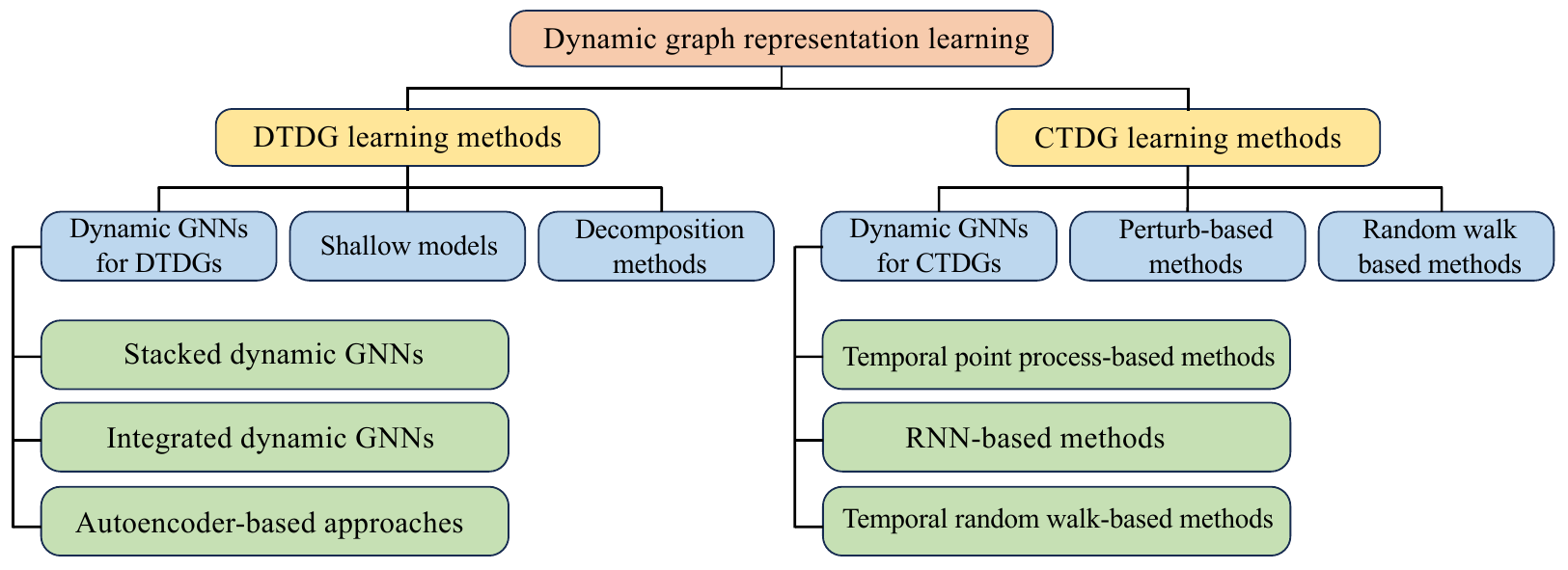} 
    \caption{\revision The taxonomy of dynamic graph representation learning.}
    \label{fig:taxonomy} 
\end{figure*}

The nodes and edges of dynamic graphs may change over time, making traditional static GNNs difficult to apply directly, which introduces new challenges for the analysis and learning of dynamic graphs. Recently, researchers have integrated GNNs with sequence learning to develop dynamic GNN models, such as the well-known TGAT~\cite{xu2020tgat}, TGN~\cite{rossi2020tgn} and ROLAND~\cite{you2022roland}. These models aggregate features from neighboring nodes and incorporate a time series module, enabling the modeling of both structural features and temporal dependencies within dynamic graphs. As a result, dynamic GNNs can generate evolving node representations that accurately capture the progression of time.

Although some work has surveyed methods for dynamic graph representation learning~\cite{survey2020representation, survey2022encoder, survey2021embedding}, with the continuous emergence of new methods and applications, the content of existing reviews has become outdated and cannot reflect the latest research developments and technological trends. Therefore, this paper offers a comprehensive review of recent developments in dynamic GNN models, encompassing representation and modeling techniques for dynamic graphs, along with a comparison of existing models. The taxonomy of existing methods is illustrated in Figure~\ref{fig:taxonomy}. We also explore new research directions, including large-scale applications and pre-training strategies, summarize the current state of research, and present an outlook on future trends.

Influenced by the available public datasets, node classification and link prediction are widely used as evaluation tasks for dynamic GNNs in current research. They are extensions of traditional static graph learning tasks to dynamic graphs, where features and labels may involve changes in the temporal dimension. Some studies also focus on domain-specific graph data to accomplish tasks such as knowledge graph completion~\cite{cai2022temporalkg}, stock price prediction~\cite{liu2023stock}, and traffic flow prediction~\cite{wang2020traffic}. It is important to note that there exists a category of dynamic graphs in literature where the graph structure remains static, but the attributes of the nodes change over time. Such graphs are referred to as spatio-temporal graphs, which are considered to be outside of the scope of this survey. The main contributions are summarized as follows:
\begin{itemize}
\item We provide a comprehensive review of the fundamental concepts and characteristics of dynamic GNNs, construct the knowledge system of dynamic GNNs, and systematically investigate the key techniques and models.
\item We present in detail the current mainstream dynamic GNN models and analyze in depth their advantages and disadvantages.
\item We discuss the challenges facing this field, highlighting potential future research directions and technological trends.
\end{itemize}

{\revision The remainder of this paper is organized as follows. In Section~\ref{sec:pre} we begin with important definitions and background knowledge about dynamic graphs. The problem statement of dynamic graph representation learning is explicated in Section~\ref{sec:methods_review}, which also provides a comprehensive review of dynamic GNNs following the taxonomy demonstrated in Figure~\ref{fig:taxonomy}. Section~\ref{sec:scalable} focuses on models and systems designed for large-scale dynamic graphs. Section~\ref{sec:dataset} summarizes the commonly used datasets, prediction tasks, and benchmarks. We find that new techniques such as transfer learning and pretrained technique are recently being applied to deal with the problem of dynamic graph learning, and we briefly review them in Section~\ref{sec:new_techniques}. We then discuss possible future directions in Section~\ref{sec:challenge} and conclude the paper in Section~\ref{sec:conclusion}.}

\section{Notation and Background}
\label{sec:pre}
\subsection{Notation}
Table~\ref{tab:notation} summarizes the necessary notations used in this paper. We consider the undirected graph $G=(V, E)$, where $V$ and $E$ are the vertex and edge set, respectively. The node feature matrix is denoted as $\X \in \mathbb{R}^{n \times d}$, where $n$ is the number of nodes in the graph, and $d$ is the dimension of the feature vector of each node. The connection between nodes in the graph is represented by the adjacency matrix $\A$. $\D$ is the diagonal degree matrix, and $\P = \Dn \A \Dn$ denotes the normalized adjacency matrix, with $\L = \I - \P$ representing the Laplacian matrix. $\sigma$ indicates the nonlinear activation function.

% \begin{table}[t]
% \caption{Statistics of the datasets.}
%   \label{tab:dataset}
% \vspace{-3mm}
% \begin{tabular}{ll}
% \hline
% Notation & Description \\ \hline
% G & 172             \\
% \hline
% \end{tabular}
% \end{table}

\begin{table}[t]
\caption{Notations and the corresponding definitions.}
\label{tab:notation}
\begin{tabular}{ll}
\hline
Notation                         & Description                       \\ \hline
$G$                              & the graph                         \\
$V$                              & the vertex set                    \\
$E$                              & the edge set                      \\
$n$                              & the number of nodes               \\
$\X \in \mathbb{R}^{n \times d}$ & the feature matrix                \\
$\x_i \in \mathbb{R}^d$          & the feature vector of node $i$    \\
% $\bm{x}$ & the feature vector of node $i$ \\
$\A \in \mathbb{R}^{n \times n}$ & the adjacent matrix               \\
$\D \in \mathbb{R}^{n \times n}$ & the degree matrix                 \\
% $\P = \Dn \A \Dn$                & the normalized adjacent matrix    \\
$\P \in \mathbb{R}^{n \times n}$ & the normalized adjacent matrix    \\
$\L \in \mathbb{R}^{n \times n}$ & the Laplacian matrix              \\
$\sigma$                         & the nonlinear activation function \\
$\mathscr{G}$                      & the dynamic graph \\
\hline
\end{tabular}
\end{table}

\subsection{Dynamic Graphs}
Dynamic graphs refer to intricate graph structures that involve temporal evolution. In these graphs, the nodes, edges, and properties within the graph demonstrate a state of continuous change. Dynamic graphs are commonly observed in various domains, including social networks~\cite{gao2023rumor}, citation networks~\cite{hu2021ogb}, biological networks~\cite{fu2022dppin}, and so on. Taking social networks as an example, the registration of a new user results in the addition of new nodes to the graph, while the deactivation of existing users leads to a decrease in nodes. Moreover, when a user follows or unfollows another user, it represents the creation or disappearance of edges in the graph. Additionally, attributes like a user's age, location, and hobbies can change over time, indicating that node attributes are also dynamically evolving. Based on the granularity of time step, dynamic graphs can be categorized into Discrete-Time Dynamic Graphs (DTDGs) and Continuous-Time Dynamic Graphs (CTDGs).

\begin{definition}[Discrete-Time Dynamic Graphs]
\label{def:dtdg}
Taking snapshots of the dynamic graph at equal intervals results in a discrete sequence of network evolution, defined as $\mathscr{G}=\{G^0, G^1, \cdots, G^T\}$, where $G^t=(V^t, E^t)$ $(0\leq t \leq T)$ is the $t$-th snapshot.
\end{definition}
\begin{figure}[t]
\setlength{\belowcaptionskip}{-4mm}
    \centering
    \includegraphics[width=1\linewidth]{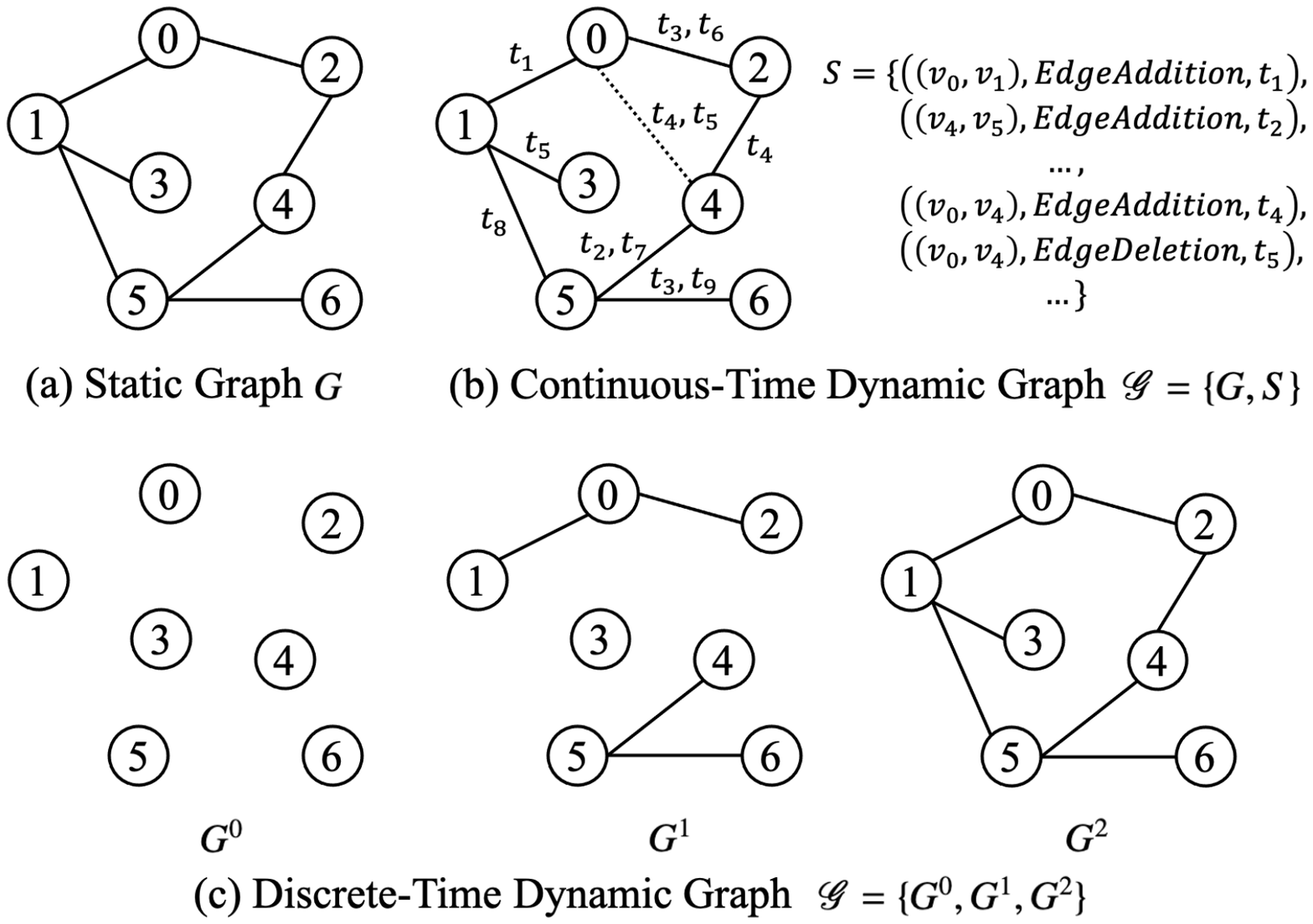} 
    % \vspace{-2mm}
    \caption{Comparative illustration of different graph representations. (a) Representation of a static graph where the structure remains unchanged over time, without temporal information; (b) Illustration of a Continuous-Time Dynamic Graph (CTDG) where interactions between nodes are labeled by timestamps, and multiple interactions are allowed; (c) Representation of a Discrete-Time Dynamic Graph (DTDG), which captures the evolution of relationships in discrete time intervals.}
    \label{fig:dynamic_graphs} 
\end{figure}

As shown in Figure~\ref{fig:dynamic_graphs}(c), DTDGs decompose the dynamic graph into a series of static networks, capturing the graph structure at various selected time points. However, some intricate details might be lost within the chosen intervals. Moreover, determining time intervals that balance accuracy and efficiency presents a significant challenge.

\begin{definition}[Continuous-Time Dynamic Graphs]
\label{def:ctdg}
A CTDG is typically defined as $\mathscr{G}=\{G, S\}$, where $G$ denotes the initial state of the graph at time $t_0$. This initial state might be empty or include initial structures (associated edges) but does not contain any history of graph events. The set $S=\{event^1, event^2, \cdots, event^T\}$ represents a collection of graph events. Each graph event is defined by a triple, which consists of the nodes participating in the event, its associated event type, and a timestamp. The event type signifies the various possible modifications occurring within a graph, including the addition or deletion of edges or nodes, as well as updates to node features. Fundamentally, these modifications may result in changes to the structure or attributes of the graph.
\end{definition}

As shown in Figure~\ref{fig:dynamic_graphs}(b), the initial graph $G$ is gradually updated according to the guidance of $S$. The structure of the graph at time $t = t_9$ is consistent with Figure~\ref{fig:dynamic_graphs}(a), and the complete process of change is recorded in $S$. It should be noted that an edge previously existed between node $v_0$ and node $v_4$, which quickly disappeared. However, the DTDG illustrated in Figure~\ref{fig:dynamic_graphs}(c) does not capture this dynamic.

\noindent \textbf{Remark.} To ensure clarity and consistency in this paper, unless otherwise specified, the graph at time $t$ refers to the $t$-th version of the graph. This could mean either the $t$-th graph snapshot in a DTDG or a graph updated with events up to time $t$ from its initial state in a CTDG.
% instance. It means that it could be either the $t$-th graph snapshot in a DTDG or a graph updated with events up to time $t$ from its initial state in a CTDG.

\vspace{-2mm}
\begin{definition}[Temporal Neighbors]
\label{def:neighbors} At time $t$, the temporal neighbor of node $v$ is represented as $N^t(v)$. Based on the traditional neighbor set, each neighbor node $w$ in the set is associated with a timestamp $t^\prime$ to form a $(node, timestamp)$ pair. It indicates that at time $t^\prime$, node $v$ is connected to node $w$. It is required that the timestamp $t^\prime$ corresponding to a neighboring node is smaller than the current timestamp $t$. In other words, $N^t(v)$ represents the collection of adjacent nodes resulting from events occurring prior to time $t$.
\end{definition}

\vspace{-6mm}
\subsection{Hawkes Processes}
Introduced by Hawkes in 1971~\cite{hawkes1971first}, the Hawkes process is a mathematical model for time-sequenced events. Based on the assumption that past events influence current events, the Hawkes process simulates a point process with an underlying self-exciting pattern. It is appropriate for modeling the evolution of discrete sequences. Given all historical events up to time $t$, the probability of an event occurring in a short time interval $[t, t+\Delta)$ at time $t$ can be defined by the conditional intensity function of the Hawkes process, denoted as
\begin{equation}
\begin{aligned}
\lambda^*(t) &=\lim _{\Delta t \rightarrow 0} \frac{\mathbb{E}[H(t+\Delta t)-H(t) \mid \mathcal{H}(t)]}{\Delta t} \\
&=\mu+\int_0^t g(t-s) d N(s),
\end{aligned}
\end{equation}
where $\mathcal{H}(t)$ represents historical events prior to time $t$, $H(t)$ denotes the number of historical events that occurred before time $t$, and $\Delta t$ indicates the window size. The first equation describes the probability that an event will occur in the region beyond the observed data $\mathcal{H}(t)$. Specifically, it represents the probability that an event will occur at time $t + \Delta t$ based on prior observations up to time $t$. The background intensity is represented by $\mu$, and the excitation function is denoted by $g(\cdot)$. Typically, $g(\cdot)$ is defined as an exponential decay function, such as $e^{-\beta t} (\beta > 0)$, which describes the phenomenon where the influence of an event diminishes as the time interval increases. According to the second equation, $\lambda^*(\cdot)$ is a flexible function that depends only on the background intensity and the excitation function. Therefore, for an observed event sequence $\{t_1, t_2, t_3, \cdots\}$, the intensity function of the Hawkes process is defined as
\vspace{-2mm}
\begin{equation}
\lambda^*(t)=\mu+\sum_{t_i<t} \alpha_i g\left(t-t_i\right),
\vspace{-2mm}
\end{equation}
where $\alpha_i$ represents the weight of the impact of each event at time $t$.

Initially, the Hawkes process was utilized for sequence modeling~\cite{zuo2020transformer, lu2019temporal}. Following the development of graph representation learning, HTNE~\cite{zuo2018htne} first introduced the Hawkes process into dynamic graph representation learning. It models the evolutionary incentive of historical behaviors on current actions, thereby simulating the graph generation process. Lu et al.~\cite{lu2019micro} proposed a dynamic graph representation model that combines macroscopic graph evolution with microscopic node edge prediction, thereby capturing the dynamic nature of the graph structure and the rising trend of graph scale. 
% Existing methods frequently neglect the variety of event types and incentives in favor of event generation.

\section{Graph Representation Learning}
\label{sec:methods_review}
% \subsection{Models for Static Graphs}
The objective of graph representation learning is to build low-dimensional vector representations of nodes while preserving important information such as structural, evolutionary, or semantic features. This has emerged as a crucial concern in graph data mining and analysis in recent years. The learned representations have extensive applications in various tasks, including node classification, link prediction, and recommendation.

It serves as a bridge between raw graph data and graph analysis tasks by encoding each node into a vector space. This approach transforms graph learning tasks into corresponding problems in machine learning applications. For example, we can effectively and efficiently address these graph analysis challenges by using the learned representations as the input feature for traditional machine learning algorithms. Additionally, we can further enhance this process by leveraging suitable classification and clustering techniques for tasks such as node classification and community detection.

\begin{definition}[Graph Representation Learning]
\label{def:grl}
Given a graph $G = (V, E)$, the task of graph representation learning is to learn an efficient mapping function $\mathcal F(v) \rightarrow \z_v$, where $\z_v \hspace{-0.5mm}\in\hspace{-0.5mm} \mathbb{R}^d$ represents the low-dimensional representation of node $v$, and $d \ll n$, $n$ is the number of nodes in the graph.
\end{definition}
Graph representation learning has been extensively studied in recent years. Earlier methods attempted to decompose data matrices into lower-dimensional forms while preserving the manifold structures and topological properties inherent to the original matrices. Prominent examples of this approach include techniques such as Local Linear Embedding (LLE)~\cite{roweis2000lle} and Laplacian Eigenmaps (LE)~\cite{belkin2003le}. These matrix factorization-based methods were further developed to incorporate higher-order adjacency matrices, thereby preserving the structural information of graphs, such as GraRep~\cite{cao2015grarep} and HOPE~\cite{zhu2018hope}. Another strategy for graph embedding creates multiple paths by traversing from random initial nodes to capture contextual information about global and local structural information of neighboring nodes, namely random walk-based methods. Probabilistic models such as Skip-gram~\cite{bartunov2016skipgram} and Bag-of-Words~\cite{joachims1998bagofwords} are then employed to sample these paths and learn node representations randomly, and typical examples include DeepWalk~\cite{perozzi2014deepwalk} and Node2vec~\cite{grover2016node2vec}. The success of deep learning techniques has paved the way for an abundance of deep neural network-based graph representation learning techniques. Graph embedding methods based on deep learning typically apply deep neural networks directly to the entire graph (or its adjacency matrix), encoding it into a lower-dimensional space. In this field, notable methods include SDNE~\cite{wang2016sdne}, DNGR~\cite{cao2016dngr}, ChebNet~\cite{defferrard2016chebnet}, GCN~\cite{kipf2016gcn}, GAT~\cite{velivckovic2018gat}, and GraphSAGE~\cite{hamilton2017graphsage}.

Unlike traditional static graphs, which focus solely on maintaining structural features, dynamic graphs also incorporate evolutionary features. Dynamic graph representation learning employs graph representation techniques for dynamic graphs and tracks the evolution of node representations. As the graph evolves over time, the representations of the nodes adapt accordingly. A time-dependent function expresses this evolution, and we denote the updated representation vector following an event involving node $v$ at time $t$ as $\z_v^t$.
\begin{definition}[Dynamic Graph Representation Learning]
\label{def:dgrl}
Given a dynamic graph $\mathscr{G}$ evolving over time, the task of dynamic graph representation learning is to learn an efficient mapping function $\mathcal{F}(v, t) \rightarrow \z_v^t$, where $\z_v^t \in \mathbb{R}^d$ represents the time-dependent low-dimensional representation of node $v$ at time $t$. Here, $d \ll n^t$, and $n^t$ is the number of nodes in the graph at time $t$.
\end{definition}

\subsection{Models for Dynamic graphs}
Corresponding to the different types of dynamic graphs described in Definitions~\ref{def:dtdg} and~\ref{def:ctdg}, dynamic graph representation learning models can be divided into two main categories: discrete-time dynamic graph representation learning and continuous-time dynamic graph representation learning. The former divides the dynamic graph into multiple snapshots, which can be regarded as multiple static graphs. Consequently, methods designed for static graphs can be used to learn representations for each snapshot, while also analyzing interrelationships and evolutionary trends between successive snapshots. In continuous-time dynamics, interactions between graph nodes are tagged with timestamps, effectively differentiating each edge based on its timestamp. Continuous-time dynamic graph representation learning refers to the methods that utilize this edge information with accompanying timestamps for learning representations.

By introducing matrix perturbation theory, the matrix decomposition-based representation learning method designed for static graphs can be extended to the learning of dynamic graphs. A few assumptions and operations form the basis for matrix perturbation theory. First, it is assumed that the adjacency matrix of the current time step is a perturbed version of the matrix from the previous time step. Then, a common low-rank decomposition is performed on these two matrices, ensuring that the representation matrix of the current time step is updated smoothly from the previous time step. By constraining the magnitude of the perturbation, it is possible to regulate the smoothness of the representation between time steps~\cite{zhu2018dhpe, li2017dane}. However, this method is more suited to dealing with DTDGs than CTDGs, as it may face challenges such as high computational complexity, difficulties in real-time updates, and inadequacies in capturing temporal dependencies on frequently changing CTDGs.

By requiring that random walks on graphs strictly follow the order of timestamps, it is possible to perform temporal walks on dynamic graphs, particularly suitable for CTDGs. For example, CTDNE~\cite{nguyen2018ctdne} incorporates a biased sampling mechanism that exhibits temporal information during the sampling process, ensuring a higher probability of sampling the edges with smaller time intervals. EvoNRL~\cite{goyal2020evonrl}, on the other hand, designs strategies for four distinct scenarios in dynamic graphs, i.e., the addition and removal of edges or nodes, updating node representations with each change of the graph.

Traditional methods rely more frequently on manually crafted features and customized models, resulting in limited learning and adaptability. Rooted in deep learning principles, GNNs have shown a remarkable ability to learn complex patterns and features in static graphs, exhibiting improved representation learning performance as well as superior generalization abilities. As a result, there has been a growing interest in utilizing GNN-inspired methods in the field of dynamic graph learning. These approaches leverage the powerful learning capabilities of GNNs to enhance the comprehension of dynamic graphs that exhibit complex temporal evolution patterns, which are known as Dynamic Graph Neural Networks (Dynamic GNNs). Recently, there has been an explosion of work on the design of Dynamic GNNs. Therefore, we introduce Dynamic GNNs that deal with DTDGs and CTDGs in Sections~\ref{sec:gnns_dtdg} and~\ref{sec:gnns_ctdg}, respectively. These two types of methods capture temporal evolution patterns in various dynamic graphs and have achieved excellent performance in several dynamic graph analysis tasks. To highlight the strengths and weaknesses of each method, we provide an in-depth comparison in Table~\ref{tab:methods_comparison}, focusing on the base models used, the types of graphs supported, and the graph events handled.

\begin{figure*}[t]
\setlength{\abovecaptionskip}{1mm}
\setlength{\belowcaptionskip}{-4mm}
	\begin{small}
		\centering
		\vspace{-2mm}
		%    \begin{footnotesize}
		\begin{tabular}{ccc}
			 \includegraphics[height=60mm]{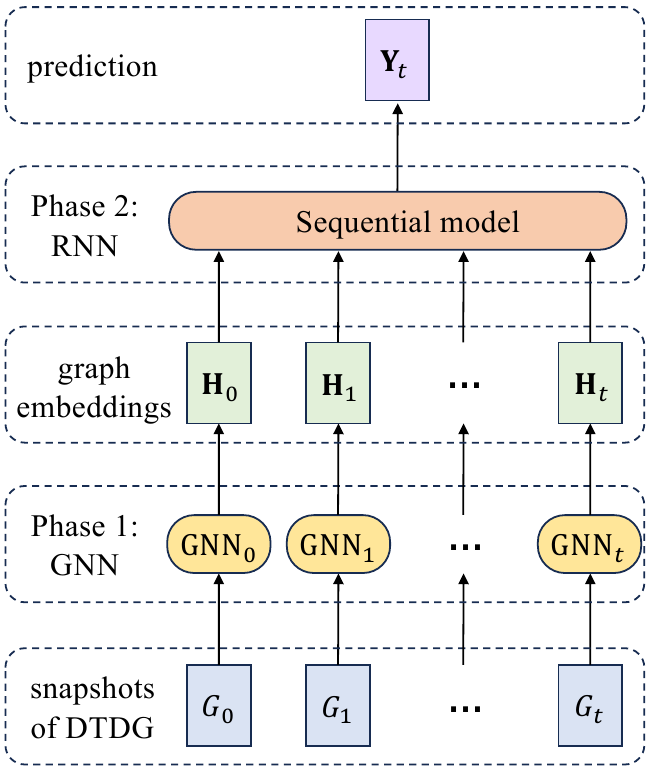} &
			 \includegraphics[height=60mm]{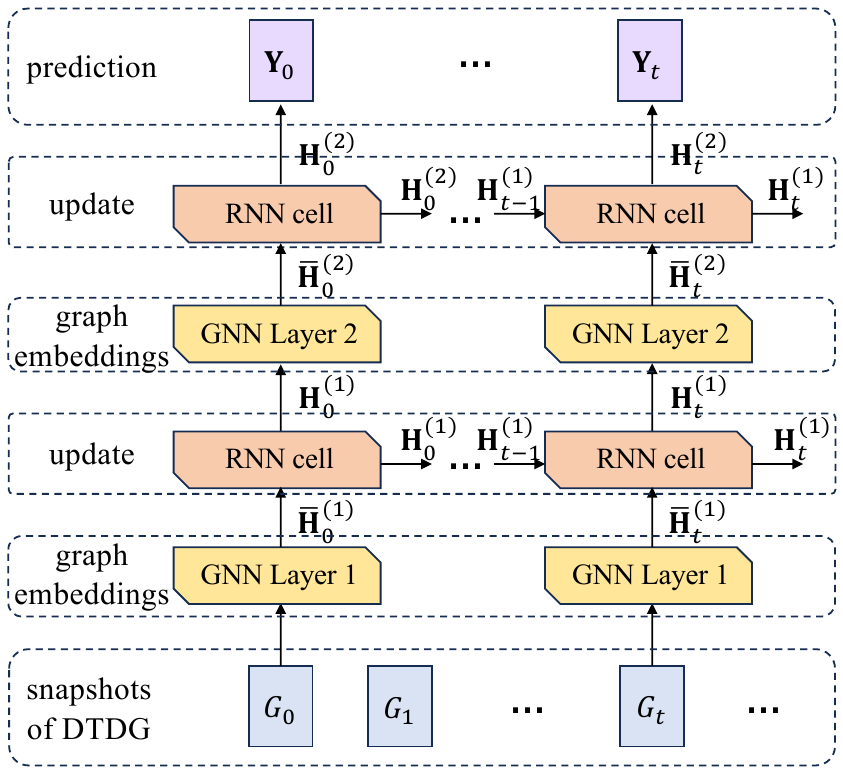} &
			 \includegraphics[height=60mm]{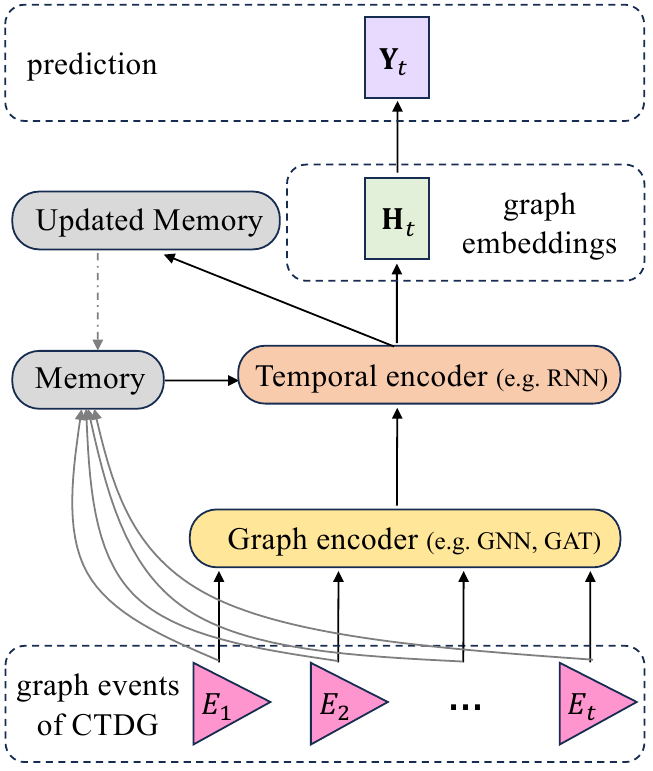} \\
              { (a) Stacked dynamic GNN for DTDGs.} & {(b) Integrated dynamic GNN for DTDGs.} & {(c) Dynamic GNN for CTDGs.} \\
		\end{tabular}
		% \vspace{-2mm}
		\caption{\revision Different model architectures for dynamic graphs.}
		\label{fig:model_archi}
% 		\vspace{-2mm}
	\end{small}
\end{figure*}
\begin{table*}[t]
\revision{
\caption{\revision Comparison of graph neural network methods in literature review. \ding{51} represents that the method supports the corresponding graph event, \ding{53} means it does not support that event, and \Square \textsuperscript{*} denotes that it considers only static attributes.}
\label{tab:methods_comparison}
\small{
\begin{tabular}{lllcccccc}
\hline
% \multirow{2}{*}{Method} & \multirow{2}{*}{Models}                      & \multicolumn{1}{l|}{\multirow{2}{*}{Graph Type}} & \multicolumn{2}{l|}{Insertion}                          & \multicolumn{2}{l|}{Deletion}                           & \multicolumn{2}{l}{Attribute}                           \\ \cline{4-9} 
%                         &                                              & \multicolumn{1}{l|}{}                            & \multicolumn{1}{l|}{Node}  & \multicolumn{1}{l|}{Edge}  & \multicolumn{1}{l|}{Node}  & \multicolumn{1}{l|}{Edge}  & \multicolumn{1}{l|}{Node}  & Edge                       \\ \hline
\multirow{2}{*}{Method} & \multirow{2}{*}{Models}                      & \multirow{2}{*}{Graph} & \multicolumn{2}{l}{Insertion}                           & \multicolumn{2}{l}{Deletion}                            & \multicolumn{2}{l}{Attribute}                           \\ \cline{4-9} 
                        &                                              &                             & \multicolumn{1}{l|}{Node}  & \multicolumn{1}{l|}{Edge}  & \multicolumn{1}{l|}{Node}  & \multicolumn{1}{l|}{Edge}  & \multicolumn{1}{l|}{Node}  & Edge                       \\ \hline
% CD-GCN~\cite{manessi2020wdgcn}      & GCN and LSTM                                 & DTDG       & \ding{53} & \ding{51} & \ding{53} & \ding{51} & \ding{51} & \ding{51} \\
WD-GCN~\cite{manessi2020wdgcn}       & GCN with skip connections and LSTM           & DTDG       & \ding{53} & \ding{51} & \ding{53} & \ding{51} & \ding{51} & \ding{51} \\
DySAT~\cite{sankar2020dysat}        & GAT and Transformer                          & DTDG       & \ding{51} & \ding{51} & \ding{51} & \ding{51} & \ding{51} & \ding{51} \\
TNDCN~\cite{wang2020tndcn}        & GNNs and spatio-temporal convolutions        & DTDG       & \ding{53} & \ding{51} & \ding{53} & \ding{51} & \ding{51} & \ding{51} \\
TEDIC~\cite{wang2021tedic}        & GNNs and spatio-temporal convolutions        & DTDG       & \ding{53} & \ding{51} & \ding{53} & \ding{51} & \ding{51} & \ding{51} \\
GC-LSTM~\cite{chen2022gclstm}      & GCN and LSTM                                 & DTDG       & \ding{53} & \ding{51} & \ding{53} & \ding{51} & \ding{51} & \ding{51} \\
LRGCN~\cite{li2019lrgcn}        & R-GCN and LSTM                               & DTDG       & \ding{53} & \ding{51} & \ding{53} & \ding{51} & \ding{51} & \ding{51} \\
RE-Net~\cite{jin2020renet}       & integrate R-GCN within RNNs                  & DTDG       & \ding{53} & \ding{51} & \ding{53} & \ding{51} & \ding{53} & \ding{53} \\
WinGNN~\cite{zhu2023wingnn}       & GNN and randomized sliding-window            & DTDG       & \ding{51} & \ding{51} & \ding{51} & \ding{51} & \Square \textsuperscript{*}    & \ding{53} \\
ROLAND~\cite{you2022roland}       & GNNs and RNNs                                & DTDG       & \ding{51} & \ding{51} & \ding{51} & \ding{51} & \ding{51} & \ding{51} \\
EvolveGCN~\cite{pareja2020evolvegcn}    & GCN and LSTM/GRU                             & DTDG       & \ding{51} & \ding{51} & \ding{51} & \ding{51} & \ding{51} & \ding{51} \\
SEIGN~\cite{qin2023seign}        & GCN and LSTM/GRU                             & DTDG       & \ding{51} & \ding{51} & \ding{51} & \ding{51} & \ding{51} & \ding{51} \\
dyngraph2vec~\cite{goyal2020dyngraph2vec} & AE and LSTM                                  & DTDG       & \ding{51} & \ding{51} & \ding{51} & \ding{51} & \ding{53} & \ding{53} \\
\hline
Know-Evolve~\cite{trivedi2017knowevolve}  & RNN-parameterized TPP                        & CTDG       & \ding{53} & \ding{51} & \ding{53} & \ding{53} & \ding{53} & \ding{53} \\
DyREP~\cite{trivedi2019dyrep}        & RNN-parameterized TPP                        & CTDG       & \ding{51} & \ding{51} & \ding{53} & \ding{53} & \ding{53} & \ding{53} \\
LDG~\cite{knyazev2021ldg}          & RNN-parameterized TPP and self attention     & CTDG       & \ding{51} & \ding{51} & \ding{53} & \ding{53} & \ding{53} & \ding{53} \\
GHNN~\cite{han2020ghnn}         & LSTM-parameterized TPP                       & CTDG       & \ding{53} & \ding{51} & \ding{53} & \ding{53} & \ding{53} & \ding{53} \\
GHT~\cite{sun2022ghtrans}          & LSTM-parameterized TPP                       & CTDG       & \ding{53} & \ding{51} & \ding{53} & \ding{53} & \ding{53} & \ding{53} \\
TREND~\cite{wen2022trend}        & Hawkes process-based GNN                     & CTDG       & \ding{51} & \ding{51} & \ding{53} & \ding{53} & \Square \textsuperscript{*}    & \ding{53} \\
DGNN~\cite{ma2020dgnn}         & GNNs and LSTM                                & CTDG       & \ding{51} & \ding{51} & \ding{53} & \ding{53} & \ding{53} & \ding{53} \\
JODIE~\cite{kumar2019jodie}        & GNNs and RNNs                                & CTDG       & \ding{51} & \ding{51} & \ding{53} & \ding{53} & \Square \textsuperscript{*}    & \ding{51} \\
TGAT~\cite{xu2020tgat}         & GAT, time encode and RNNs                    & CTDG       & \ding{51} & \ding{51} & \ding{53} & \ding{53} & \Square \textsuperscript{*}    & \ding{51} \\
TGN~\cite{rossi2020tgn}          & GNNs, time encode and RNNs                   & CTDG       & \ding{51} & \ding{51} & \ding{53} & \ding{53} & \Square \textsuperscript{*}    & \ding{51} \\
APAN~\cite{wang2021apan}         & GNNs and RNNs                                & CTDG       & \ding{51} & \ding{51} & \ding{53} & \ding{53} & \Square \textsuperscript{*}    & \ding{51} \\
CAW-N~\cite{wang2020caw}        & causal anonymous walk based GNN              & CTDG       & \ding{51} & \ding{51} & \ding{53} & \ding{53} & \ding{53} & \ding{51} \\
Zebra~\cite{li2023zebra}        & temporal PPR based GNN                       & CTDG       & \ding{51} & \ding{51} & \ding{53} & \ding{53} & \ding{51} & \ding{51} \\ 
EARLY~\cite{li2023early}        & GCN and temporal random walk         & CTDG          & \ding{51} & \ding{51} & \ding{51} & \ding{51} & \ding{51} & \ding{51} \\
Zheng et al.~\cite{zheng2023decoupled} & decoupled GNNs and sequential models & CTDG & \ding{51} & \ding{51} & \ding{51} & \ding{51} & \ding{51} & \ding{51} \\
SDG~\cite{fu2021sdg}          & APPNP and dynamic propagation       & CTDG          & \ding{53} & \ding{51} & \ding{53} & \ding{51} & \ding{51} & \ding{51} \\
\hline
\end{tabular}
}
% \begin{tablenotes}
%     \item \Square \textsuperscript{*} Indicates that the model considers only static attributes.
% \end{tablenotes}
}
\end{table*}

\subsection{Dynamic GNNs for DTDGs}
\label{sec:gnns_dtdg}
% \subsection{Methods Based on Variational Autoencoders}
DTDGs are composed of multiple snapshots that are chronologically organized and can be modeled as sequential data, as described in Definition~\ref{def:dtdg}. Therefore, temporal patterns in DTDGs are measured by the sequential relationships between different snapshots. The prevailing research methodology involves employing static graph neural network models to represent each graph snapshot individually. These snapshots are subsequently organized chronologically and treated as sequential data, which can be input into sequence models to comprehensively determine the interplay between different graph snapshots and learn temporal patterns. As a representative sequence model, Recurrent Neural Networks (RNNs) are often combined with GNNs to form the mainstream dynamic GNNs for DTDGs. Based on how they are combined, they can be broadly classified as stacked architectures or integrated architectures.

% \noindent \textbf{Stacked dynamic GNNs} combine spatial GNNs and temporal RNNs modularly to model discrete sequences of dynamic graphs. {\revision As shown in Figure~\ref{fig:model_archi}(a), the distinct components are dedicated to spatial and temporal modeling, respectively.} Specifically, these models use distinct GNNs to aggregate spatial information on each graph snapshot individually. The outputs from the different time steps are then fed into a temporal module for sequence modeling. Therefore, given a graph $G^t$ at time $t$, the representation is computed as follows:
\noindent \textbf{Stacked dynamic GNNs} combine spatial GNNs with temporal RNNs in a modular fashion to model discrete sequences of dynamic graphs. As shown in Figure~\ref{fig:model_archi}(a), the distinct components are dedicated to spatial and temporal modeling, respectively. Specifically, these models use distinct GNNs to aggregate spatial information on each graph snapshot individually. The outputs from the different time steps are then fed into a temporal module for sequence modeling. Therefore, given a graph $G^t$ at time $t$, the representation is computed as follows:
\begin{equation}
\Z^t = f\left(G^t\right), \H^t = g\left(\H^{t-1}, \Z^t\right),
\end{equation}
where $f$ is a GNN model used to encode $G^t$ and obtain the graph representation $\Z^t$ at this time step. $g$ is a sequence model such as RNN, used to integrate the previous hidden state $\H^{t-1}$ and current graph representation $\Z^t$ to update the current hidden state $\H^t$. Manessi et al.~\cite{manessi2020wdgcn} use this architecture to propose CD-GCN and WD-GCN models, where $f$ is a GCN model and $g$ uses a standard Long Short-Term Memory (LSTM). Notably, each node utilizes a distinct LSTM, though the weights can be shared. The difference between WD-GCN and CD-GCN is that CD-GCN adds skip connections to GCN.

RNNs are not the only available option for learning time series. In recent studies, GNNs have also been used in combination with other kinds of deep time series models. For instance, DySAT~\cite{sankar2020dysat} introduces a stacked architecture made up entirely of self-attention blocks. They use attention mechanisms in both the spatial and temporal dimensions, employing Graph Attention Networks (GAT)~\cite{velivckovic2018gat} for the spatial dimension and Transformers~\cite{zuo2020transformer} for the temporal dimension. Similarly, models like TNDCN~\cite{wang2020tndcn} and TEDIC~\cite{wang2021tedic} combine GNNs with one-dimensional spatio-temporal convolutions.

\noindent \textbf{Integrated dynamic GNNs} merge GNNs and RNNs within a single network layer, thus amalgamating spatial and temporal modeling within the same architectural level, {\revision as shown in Figure~\ref{fig:model_archi}(b)}. GC-LSTM~\cite{chen2022gclstm} uses the adjacency matrix of a given time as input to the LSTM and performs spectral graph convolution~\cite{defferrard2016convolutional} on the hidden layer. The model takes a sequence of adjacency matrices as input and returns node representations that encode both temporal and spatial information. Within this framework, the GNN learns the topological features of the LSTM's cell and hidden states, which are used to maintain long-term relationships and extract input information, respectively. This implies that the node representations provided by the GNN are distinct from the node history captured by the LSTM. Let $f_{\i}$, $f_{\f}$, $f_{\u}$, $f_{\o}$, and $f_{\c}$ represent the five GNN models acting on the input gate, forget gate, update, output gate, and cell state, respectively. They possess the same structure but different parameters. The parameters of $f_{\i}$, $f_{\f}$, $f_{\u}$, and $f_{\o}$ are initialized by the hidden state $\H^{t-1}$ from the previous time step, while $f_{\c}$ is initialized based on the memory $\C^{t-1}$:
\begin{equation}
\begin{aligned}
\i^t &= \sigma\left(\A^t \W_{\i} + f_{\i}\left(G^{t-1}\right)+\b_{\i}\right) \\
\f^t &= \sigma\left(\A^t \W_{\f} + f_{\f}\left(G^{t-1}\right)+\b_{\f}\right) \\
\C^t &= \f^t \otimes f_{\c}\left(G^{t-1}\right)+\i^t \odot \tanh \left(\A^t \W_{\c} + f_{\u} \left(G^{t-1}\right)+\b_{\c}\right) \\
\boldsymbol{o}^t &= \sigma\left(\A^t \W_{\o}+f_{\o}\left(G^{t-1}\right)+\b_{\o}\right) \\
\H^t &= \boldsymbol{o}^t \odot \tanh \left(\C^t\right),
\end{aligned}
\end{equation}
where $\{\W_{\i}, \W_{\f}, \W_{\o}, \W_{\c}\}$ and $\{\b_{\i}, \b_{\f}, \b_{\c}, \b_{\o}\}$ are the weight and bias matrices, respectively, and $\A^t$ is the adjacency matrix of $G^t$. Other integrated approaches often follow a similar framework, differing perhaps in the GNN or RNN models they employ, the use cases, or the types of graphs they focus on. The Long Short-Term Memory R-GCN (LRGCN)~\cite{li2019lrgcn} adopts a similar strategy, incorporating topological variations directly into its computations. Specifically, LRGCN~\cite{li2019lrgcn} integrates a Relational GCN (R-GCN)~\cite{li2019lrgcn} within the LSTM framework. The computations for the input gate, forget gate and output gate are the results of the R-GCN model acting on the input node representations and embeddings computed from the previous step. Meanwhile, RE-Net~\cite{jin2020renet} integrates R-GCN~\cite{li2019lrgcn} into multiple RNNs, enabling it to learn from dynamic knowledge graphs. WinGNN~\cite{zhu2023wingnn} utilizes the idea of randomized gradient aggregation over sliding windows to model temporal patterns without extra-temporal encoders. Specifically, WinGNN introduces a stochastic gradient aggregation mechanism based on sliding windows, gathering gradients across multiple consecutive snapshots to attain a global optimum. In addition, an adaptive gradient aggregation strategy is designed to assign different weights to gradients from different snapshots, overcoming the influence of local optima. It also introduces a snapshot random dropout mechanism to enhance the robustness of gradient aggregation further.

ROLAND~\cite{you2022roland} extends the GNN-RNN framework by utilizing hierarchical node states. Specifically, You et al.~\cite{you2022roland} propose to stack multiple GNN layers and interleave them with a sequence model such as RNN, where the node embedding of different GNN layers is regarded as hierarchical node states that are continuously updated over time. In contrast to traditional GNNs, ROLAND uses not only the embedding information $\H_t^{(\ell-1)}$ from the previous layer but also the information $\H_{t-1}^{(\ell)}$ from the previous moment of the current layer, when updating the embedding matrix $\H_t^{(\ell)}$ at time $t$. Thus, the $(\ell)$-th layer of the framework is
\begin{equation}
\overline{\H}_t^{(\ell)} = f^{(\ell)}\left(\overline{\H}_t^{(\ell-1)}\right), \H_t^{(\ell)} = g\left(\H_{t-1}^{(\ell)}, \overline{\H}_t^{(\ell)}\right),
\end{equation}
where $f^{(\ell)}$ is the $(\ell)$-th GNN layer, $g$ is the sequence encoder. Bonner et al.~\cite{bonner2019tna} propose stacking a GCN layer, a Gated Recurrent Unit (GRU) layer~\cite{chung2014gru}, and a linear layer to form a Temporal Neighborhood Aggregation (TNA) layer, and employ a two-layer model to implement 2-hop convolution combined with variational sampling for link prediction.

Instead of updating node representations with RNNs as in previous methods, EvolveGCN~\cite{pareja2020evolvegcn} uses RNNs to update the weight parameters of GCNs between graph snapshots, facilitating dynamic adaptation and model performance improvement. Building on this, they introduced two variants: EvolveGCN-H, employing a GRU to enhance temporal stability, and EvolveGCN-O, using an LSTM to capture longer dependencies. To improve scalability, SEIGN~\cite{qin2023seign} focuses on evolving parameters associated with the graph filter, specifically the convolutional kernel, rather than the entire GNN model.

\noindent \textbf{Autoencoder-based approaches} are trained distinctively compared to the dynamic graph neural network models discussed above. They leverage dynamic graph embeddings to capture both the topology and its temporal changes. DynGEM~\cite{goyal2018dyngem}, for instance, employs an autoencoder to learn the embeddings from each snapshot of the DTDG. Interestingly, the autoencoder at time $t$ initializes its parameters based on the autoencoder from time $t-1$. The architecture of this autoencoder, including the number of layers and neurons in each layer, is adjusted based on the differences between the current snapshot and its predecessor. The Net2Net~\cite{chen2015net2net} technique is utilized to maintain the mapping relationships across layers when modifications are made to the autoencoder. To effectively capture the relationships and changes within dynamic graphs, dyngraph2vec~\cite{goyal2020dyngraph2vec} employs a series of adjacency matrices $\{\A^{t-w+1}, \cdots, \A^{t-1}\}$ within a time window as input, where $w$ represents the size of the time window. An autoencoder is then utilized to obtain the adjacency matrix $\A^t$ for the current timestep $t$.

\subsection{Dynamic GNNs for CTDGs}
\label{sec:gnns_ctdg}
% \subsection{Methods Based on Variational Autoencoders}
As defined in Definition~\ref{def:ctdg}, CTDGs are not stored as snapshots with fixed intervals, but each event is recorded as a triplet (node, event type, timestamp), e.g., $((u,v), EdgeAddition, t)$ denotes an edge insertion event between nodes $u$ and $v$ at time $t$. Therefore, each event in a continuous dynamic graph has a timestamp, which smoothly tracks the evolution of the graph structure and allows for more accurate analysis of temporal relations and dependencies between events. In addition, discrete graphs synchronously aggregate time-slice information and cannot distinguish the sequence of events, whereas continuous graphs specify the temporal order of events and support modeling of asynchronous time intervals. Approaches to learning continuous dynamic graphs revolve around how to effectively capture and characterize the evolution of graph structure over time, {\revision as shown in Figure~\ref{fig:model_archi}(c)}. There are currently three classes of approaches to continuous-type dynamic graph neural networks: (1) temporal point process-based approaches, in which the temporal point process is parameterized by a neural network; (2) RNN-based approaches, in which node representations are maintained by an RNN-based architecture; and (3) temporal random walk-based approaches, whose essence is to capture both temporal and structural properties of graphs through temporal random walks.
% (1) RNN-based approaches, in which the node representations are maintained by an RNN-based architecture; (2) temporal point process-based approaches, in which the temporal point process is parameterized by the neural network; and (3) temporal embedding approaches, in which a temporal embedding is achieved using a Temporal Convolutional Network (TCN) in conjunction with a self-attention are combined to learn local and global temporal patterns.

\noindent \textbf{Temporal point process-based methods.}
Know-Evolve~\cite{trivedi2017knowevolve} models knowledge graphs as interactive networks, where temporal point processes are used to model the occurrence of events. It defines a bilinear relationship function for representing multi-relational interactions between nodes. The central tenet of the method is to model the occurrence of facts as a multivariate temporal point process, whose conditional intensity function is modulated by the score of the relevant relationship. This relationship score is derived from embeddings of nodes that evolve dynamically. In addition, two custom RNNs are used to update node embeddings as new graph events occur.

Based on a similar strategy, Trivedi et al. proposed DyRep~\cite{trivedi2019dyrep}, a method for modeling multiple types of graphs. DyRep develops a pairwise intensity function that quantifies the strength of relationships between nodes. It further employs self-attention mechanisms to construct a temporal point process, which enables the model to prioritize relevant information for the current task, filtering out irrelevant details. Therefore, the structural embedding for node $u$ is defined as: $\h_{\text {struct }}^{\bar{t}}(u)=\max \left(\left\{\sigma\left(q^t(u, i) \cdot \h^{\bar{t}}(i)\right), \forall i \in N^{\bar{t}}(u)\right\}\right)$, where $\h^{\bar{t}}(i)=\W_h \z^{\bar{t}}(i)+\b_h$, $\z^{\bar{t}}(i)$ is the embedding of the most recent node $i$, $\bar{t}$ represents the most recent time node $i$ was active, and $q(u, i)$ denotes the aggregation weight corresponding to the neighbor node $i$. Specifically, the model computes the relationship between node $u$ and its neighbor node $i\in N^t(u)$ by
\begin{equation}
q^t(u, i)=\frac{\exp \left(\S^t(u, i)\right)}{\sum_{i^{\prime} \in N^t(u)} \exp \left(\S^{\bar{t}}\left(u, i^{\prime}\right)\right)} ,
\end{equation}
where $\S^t \in \mathbb{R}^{n\times n}$ is the attention matrix, which represents the fraction of attention between pairs of nodes at time $t$. $\S$ is computed and maintained by the adjacency matrix $\A$ and the conditional intensity function, and the conditional intensity function between node $u$ and node $v$ is defined as
\begin{equation}
\lambda^t(u, v)=f\left(g^{\bar{t}}(u, v)\right),
\end{equation}
where $g^{\bar{t}(u, v)}$ is an internal function that computes the compatibility between two recently updated nodes: $g^{\bar{t}}(u, v)=w^{T} \cdot\left[\z^{\bar{t}}(u): \z^{\bar{t}}(v)\right]$, and $:$ denotes the concatenation, $w^{T}$ is the learnable scale-specific compatibility model parameter. $f(\cdot)$ uses the modified softplus function $f(x)=\psi \log \left(1+\exp \left(\frac{x}{\psi}\right)\right)$, where $x$ represents $g^{t}(\cdot)$, $\psi>0$ is the learnable scalar time parameter corresponding to the proportion of relevant events occurring during the process.

{\polish Latent Dynamic Graph (LDG)~\cite{knyazev2021ldg} is an enhanced version of DyREP~\cite{trivedi2019dyrep} that integrates Neural Relational Inference (NRI)~\cite{kipf2018nri} to encode temporal interactions, thereby optimizing the attention mechanism and resulting in improved performance.} Graph Hawkes Neural Network (GHNN)~\cite{han2020ghnn} and Graph Hawkes Transformer (GHT)~\cite{sun2022ghtrans} model temporal dependencies between nodes using Hawkes processes, and estimate the intensity functions using continuous-time LSTM and Transformer respectively to learn time-evolving node representations. TREND~\cite{wen2022trend} also designs a Hawkes process-based GNN to learn representations for CTDGs, where the parameters of the conditional intensity are learned through a Fully Connected Layer (FCL), and Feature-wise Linear Modulation (FiLM) is used to learn the parameters of the FCL.

\noindent \textbf{RNN-based methods.}
These models use RNNs to maintain node embeddings in a continuous manner, which allows them to respond to graph changes in real time and update the representations of affected nodes. Specifically, such models define an RNN unit to track the evolution of each node embedding. When the graph changes, the RNN unit can calculate the new hidden state at the current time $t$ based on the current input and the hidden state at the previous moment. Therefore, one of the main differences between these methods lies in how they define the embedding function and customize the RNN.

DGNN~\cite{ma2020dgnn} consists of an update component and a propagation component. When introducing a new edge, the update component maintains the freshness of node information by capturing the order of edge additions and the time interval between interactions, implemented by the LSTM model. The propagation component propagates the new interaction information to the affected nodes by considering the intensity of the influence.

JODIE~\cite{kumar2019jodie} is composed of two operations: the update operation and the projection operation. The former is used to update the embedding information of nodes, while the latter is used to predict future embedding trajectories. Moreover, each node has a static embedding to represent its fixed attributes, and a dynamic embedding to reflect its current state. JODIE focuses on interaction network data. In the update operation, the embeddings of the user node $u$ and the item node $i$ are updated by two RNNs that have the same structure but different parameters:
\begin{equation}
\begin{aligned}
\z^t(u)&=\sigma\left(\W_{u, 1} \z^{\bar{t}}(u)+\W_{u, 2} \z^{\bar{t}}(i)+\W_{u, 3} e_{u, i}+\W_{u, 4} \Delta_u\right) \\
\z^t(i)&=\sigma\left(\W_{i, 5} \z^{\bar{t}}(i)+\W_{i, 6} \z^{\bar{t}}(u)+\W_{i, 7} e_{u, i}+\W_{i, 8} \Delta_i\right),
\end{aligned}
\end{equation}
where $ \z^{\bar{t}}(u) $ and $ \z^{\bar{t}}(i) $ represent the previous embeddings of node $ u $ and node $ i $ respectively. $ \Delta_u $ and $ \Delta_i $ represent the time difference from the last interaction. $ e_{u, i} $ is the feature vector corresponding to the current interaction between node $ u $ and node $ i $. The sets $ \{\W_{u, 1}, \cdots, \W_{u, 4}\} $ and $ \{\W_{i, 5}, \cdots, \W_{i, 8}\} $ represent the parameters for the two RNN, $ RNN_u $ and $ RNN_i $, respectively. $\sigma $ is the sigmoid function.

The embedding projection operation is designed to generate the future embedding trajectory of a user, and this embedding can be applied to downstream tasks. The projection operation takes two inputs: (1) the embedding of node $u$ at time $t$, denoted as $\z^t(u)$, and (2) the time difference, $\Delta$. By projecting $\z^t(u)$, the representation of node $u$ over a certain time interval can be learned as:
\begin{equation}
\hat{\z}^{t+\Delta}(u)=\left(1+\W_p \Delta\right) * \z^t(u),
\end{equation}
where $\W_p$ is the learned matrix. As $\Delta$ increases, the offset of the projection embedding becomes larger. Assume that item $j$ interacts with user $u$ at time $t + \Delta$, the embedding of node $j$ at time $t + \Delta$ is predicted based on a fully connected linear layer:
\begin{equation}
\z^{t+\Delta}(j) \hspace{-0.5mm}=\hspace{-0.5mm} \W_1 \hat{\z}^{t+\Delta}(u)+\W_2 \bar{\z}(u)+\W_3 \z^{t+\Delta}(i)+\W_4 \bar{\z}(i)+\B,
\end{equation}
where $\bar{\z}(u)$ and $\bar{\z}(i)$ represent the static embeddings of nodes $u$ and $i$, respectively. $\W_1, \W_2, \W_3, \W_4$ and $\B$ are the parameters of this linear layer.

Inspired by Time2Vec~\cite{kazemi2019time2vec}, TGAT~\cite{xu2020tgat} introduces a time encoding function based on the Bochner theorem~\cite{loomis2013bochner}, which can map continuous time to a vector space. This method presents a viable substitute for position encoding within self-attention mechanisms, effectively leveraging temporal information. By utilizing the improved self-attention framework, TGAT~\cite{xu2020tgat} presents a temporal graph attention layer specifically designed to incorporate feature data from historical neighbors of a target node.

% TGN~\cite{rossi2020tgn} expands on prior concepts to present a unified framework composed of five core modules: 
TGN~\cite{rossi2020tgn} presents a unified approach by integrating the methodological frameworks of TGAT~\cite{xu2020tgat}, JODIE~\cite{kumar2019jodie}, and DyREP~\cite{trivedi2019dyrep}, which is composed of five core modules: Memory, Message Function, Message Aggregator, Memory Updater, and Embedding. The Memory module stores historical information about nodes, and the Embedding module produces real-time representations. Whenever an event involving a node occurs, the corresponding memory is updated. This update function can be learning models like RNNs or non-learning methods like averaging or aggregation, and thus take advantage of previous messages to refresh the memory during training, facilitating the generation of node embeddings. Furthermore, by ingeniously architecting each module, TGN can encapsulate a variety of models, including the likes of JODIE~\cite{kumar2019jodie}, DyRep~\cite{trivedi2019dyrep}, and TGAT~\cite{xu2020tgat}.

During online training, it cannot be guaranteed that graph update events arrive in timestamp order, which will make RNN-based embedding generation unstable. APAN~\cite{wang2021apan} addresses this issue by employing an asynchronous propagation mechanism. It allows messages to propagate through the graph asynchronously by specifying the number of propagation steps and thus ensures events are stored in order of their timestamps.

\noindent \textbf{Temporal random walk-based methods.}
Inspired by models such as DeepWalk~\cite{perozzi2014deepwalk} and Node2vec~\cite{grover2016node2vec}, Nguyen et al.~\cite{nguyen2018ctdne} first proposed a temporal random walk based on the properties of dynamic graphs, where a random walk consists of a sequence of edges with non-decreasing timestamps. Therefore, the movement from one vertex to another is dictated not only by the adjacency of the nodes but also by temporal constraints, ensuring that the walks align with both the graph's structure and the sequence of time.

Starting from links of interest, Causal Anonymous Walk (CAW)\cite{wang2020caw} traces back several neighboring links to encode the latent causal relationships in network dynamics. During walks, CAW\cite{wang2020caw} removes node identities to facilitate inductive learning and instead encodes relative node identities based on their frequencies in a set of sampled walks. This method of relative node identification ensures that pattern structures and their correlations are preserved after node identities are removed, achieving set-based anonymization. To predict temporal links between two nodes, the CAW-Network (CAW-N) model is proposed. It samples some CAWs associated with the two nodes, encodes and aggregates these CAWs using RNNs and set pooling, respectively, to make predictions.

Zebra~\cite{li2023zebra} introduces the Temporal Personalized PageRank (T-PPR) metric that uses an exponential decay model to estimate the influence of nodes in dynamic graphs. Given a source node $(i, \tau_1)$ and a target node $(j,\tau_2)$, where $(i,\tau_1)$ represents node $i$ at time $\tau_1$, the T-PPR value $\pi_{i,\tau_1}(j,\tau_2)$ of the target node $(j,\tau_2)$ with respect to the source node $(i, \tau_1)$ is defined as the probability of starting a temporal random walk from $(i, \tau_1)$ with a parameter $\alpha$ $(0<\alpha<1)$, \eat{applying an exponential decay model with parameter $\beta$ $(0<\beta<1)$, }and eventually stopping at $(j,\tau_2)$. T-PPR operates under the premise that recent neighboring nodes hold paramount importance. Therefore, by applying an exponential decay model parameterized by $\beta$ $(0<\beta<1)$, it ensures that the probability of traversing the most distant ones is the lowest among all the historical neighbors of a given node. For example, suppose source node $(v_1,t_6)$ has three historical neighbor nodes $(v_4,t_2)$, $(v_2,t_4)$ and $(v_3,t_5)$, assigned with transition probabilities $\beta^3$, $\beta^2$, $\beta$. According to the definition, the T-PPR value of $(v_3, t_5)$ can be computed as: $\pi_{v_1,t_6}(v_3, t_5) = \frac{\beta}{\beta+\beta^2+\beta^3} \times \alpha(1-\alpha)$. Based on the T-PPR values, Zebra~\cite{li2023zebra} can discern the top-$k$ temporal neighbors with the most significant impact on the target node. It directly aggregates these $k$ high-influence temporal neighbor nodes to generate the target node's embedding, avoiding complex recursive temporal neighbor aggregation.

\section{\revision Processing and Learning on Large-scale Dynamic Graphs}
\label{sec:scalable}
Several strategies are available for managing large-scale static graphs. First, to improve processing efficiency, distributed parallel training can be implemented by dividing the graph across multiple devices for simultaneous processing, which ensures data integrity. Second, to enhance throughput and optimize resource use, techniques such as pipeline parallelism, data parallelism, and model parallelism are widely employed. Third, to address computational complexity, sampling methods and neighbor truncation are utilized while preserving the integrity of critical data. Meanwhile, efficient communication methods, such as peer-to-peer exchanges, are often implemented to reduce data transfer times.

% Dynamic graph neural networks present unique challenges when handling large-scale graphs, such as the need for mini-batch training and processing in chronological order. For mini-batch training, the model must maintain a sufficient history horizon, optimize the node memory update mechanism, and share information using memory synchronization after partitioning into subgraphs. 
However, managing large-scale graphs becomes considerably more challenging when dealing with dynamic graphs. For example, if we employ mini-batch training techniques, the model must maintain a sufficient history horizon, optimize the memory update mechanism for nodes, and transfer information using memory synchronization after partitioning into subgraphs. For temporal sequential processing, the main approaches include using time encoding to capture temporal information, generating batches of data in chronological order, and ensuring that all data are traversed in an overall sequential manner. {\polish Collectively, these methodologies and fundamental prerequisites contribute to enhancing parallelism, scalability, and efficient processing capabilities for large dynamic graphs.

Existing approaches can be grouped into two primary lines. One approach gives precedence to scalability in the design of dynamic GNNs, resulting in various algorithms aimed at efficient learning on dynamic graphs. The second focuses on the development of scalable training frameworks that enable the adaptation of existing algorithms to large-scale dynamic graphs.}

% \noindent \textbf{Algorithms Supporting Large-scale Dynamic Graph Learning.} 
\subsection{\revision Scalable GNNs for Dynamic Graphs}
Instead of adapting training frameworks to fit existing algorithms for large-scale dynamic graph learning, several studies have suggested developing algorithms specifically designed for large-scale dynamic graphs. 
% Typically, these algorithms begin by preprocessing the graph data to construct a sequence of dynamic graph snapshots. Each graph snapshot learns node representations using graph neural networks such as GCN~\cite{kipf2016gcn} and GAT~\cite{xu2020tgat}. Node representations at different time steps are then modeled using sequence models, such as RNNs and attention mechanisms, capturing the evolution patterns of nodes. The final representations are utilized for downstream prediction tasks. 
{\polish To ensure the efficiency and scalability of the model, techniques like sampling and parallel training are often integrated for optimization.} Following this framework, SEIGN~\cite{qin2023seign} employs a parameter-free messaging strategy for preprocessing and uses the Inception architecture to obtain multi-scale node representations. Similarly to EvolveGCN~\cite{pareja2020evolvegcn}, SEIGN updates the convolutional kernel weights with GRU modules for each graph snapshot to maintain their freshness. However, instead of evolving the entire GNN model, SEIGN refines only the parameters of the convolutional kernel, thereby enhancing the method's scalability. Furthermore, SEIGN introduces the evolution of node representations, increasing the expressiveness of the final node representation. Additionally, it facilitates efficient graph mini-batch training, further enhancing the model's scalability.

EARLY~\cite{li2023early} reduces computational costs by updating only the top-$k$ nodes most susceptible to event impacts to meet the frequent changes in CTDGs. The selection of these top-$k$ nodes is guided by two primary metrics: local and global scores. Local scores emphasize the theoretical analysis of graph convolution networks, addressing both the effect of node updates and the variations in random walk probabilities. On the other hand, global scores emphasize the node influence. Ultimately, an information gain function integrates these scores to identify the most influential top-$k$ nodes.

Inspired by scalable GNNs on static graphs~\cite{wu2019sgc}, Zheng et al.~\cite{zheng2023decoupled} proposed a decoupled GNN model that separates the graph propagation process and the training process for the downstream tasks, where the unified dynamic propagation process can efficiently handle both continuous and discrete dynamic graphs. By leveraging the identity equation of the graph propagation process, the model locates affected nodes and quantifies the impact on their representations from graph changes. Other node representations are adjusted in the next step. This approach allows incremental updates from prior results, avoiding costly recalculations. Additionally, since computations related to the graph structure only occur during propagation, subsequent training for downstream tasks can be performed separately without involving expensive graph operations. Therefore, arbitrary sequence learning models can be plugged in. Similarly, SDG~\cite{fu2021sdg} employs a dynamic propagation matrix based on personalized PageRank, replacing the static probability transition matrix of APPNP~\cite{gasteiger2018appnp}. \eat{This not only ensures effective information transfer within the graph structure but also enhances the model's adaptability in dynamic settings. }Specifically, SDG first extracts hidden node features using a model-agnostic neural network. These features are then multiplied with the dynamic propagation matrix to disseminate neighborhood node information. The dynamic propagation matrix tracks the steady-state distribution of random walks following graph topology changes using the push-out and add-back algorithm, eliminating redundant propagation computations in dynamic graphs.

% \noindent \textbf{Distributed Training Frameworks.} 
% \subsection{\revision Distributed Training Frameworks}
\subsection{\revision Scalable Training Frameworks for Dynamic GNNs}
\label{sec:scalable_frameworks}
{\revision
\textbf{Frameworks designed for DTDGs.} Memory management optimization to support efficient model parallelism is a prevalent method for enhancing the training efficacy of GNNs for DTDGs. To optimize the end-to-end training performance of dynamic graph neural networks on a single GPU, PiPAD~\cite{wang2023pipad} introduces a pipeline execution framework and multi-snapshot parallel processing manner. Additionally, it employs a slice-based graph representation to efficiently extract overlapping sections of the graph, enabling parallel computations for multiple snapshots simultaneously. DGNN-Booster~\cite{chen2023dgnnbooster} uses the Field-Programmable Gate Array (FPGA) to enhance the inference speed of dynamic graph neural networks in a single-machine setting. It also introduces two distinct dataflow designs to support mainstream DGNN models.  Chakaravarthy et al.~\cite{chakaravarthy2021efficient} employed optimization techniques such as gradient checkpointing and graph-difference-based CPU-GPU data transfer to enhance the training of dynamic GNNs on single-node multi-GPU systems. Additionally, they introduced an algorithm that uses snapshot partitioning to train large-scale dynamic GNNs on distributed multi-node multi-GPU systems efficiently. Experimental results indicate that this method provides notable improvements over traditional methods.

\noindent \textbf{Frameworks designed for CTDGs.} To overcome the limitations of single-machine configurations and further improve the scalability of dynamic graph neural network models for processing large dynamic graphs, more prevalent strategies involve partitioning the graph across multiple devices for parallel training. In this context, random division and vertex-based partitioning are frequently employed. TGL~\cite{zhou2022tgl} proposes a generalized framework to train different dynamic GNNs, including snapshot-based, time encoding-based and memory-based methods. Temporal-CSR data structures are designed to enable fast access to temporal edges, while parallel samplers support efficient implementation of various temporal neighbor sampling algorithms. To overcome the timeliness issue of node memory when training with large batches, a random chunk scheduling technique is proposed, which enables efficient multi-GPU training on large-scale dynamic graphs. Numerous dynamic GNNs maintain node-level memory vectors to summarize historical information about each node. To maintain and update these node memories efficiently, TGL designs the mailbox to cache the latest messages. In each training iteration, the latest messages are read from the mailbox, aggregated by a certain combiner. Then, the memory state of the node is updated by a sequential model such as GRU. This design circumvents the direct generation of mail using the most recent memory state, effectively avoiding potential information leakage. Additionally, it incorporates an asynchronous mechanism for updating node states, making it better aligned with the streaming or batch training process.

\begin{table*}[t]
\centering
\revision{
\caption{\revision Scalable Dynamic GNNs training framework.}
\label{tab:training_frameworks}
\begin{tabular}{l l l l} 
\hline
Name & Graph Type & Hardware & Software \\ \hline
PiPAD~\cite{wang2023pipad} & DTDG & single-node single-GPU with multi-core CPU & PyG Temporal \\
DGNN-Booster~\cite{chen2023dgnnbooster} & DTDG & single-node single-GPU with multi-core CPU & Pytorch \\
Chakaravarthy et al.~\cite{chakaravarthy2021efficient} & DTDG & multi-node multi-GPU & PyTorch \\
TGL~\cite{zhou2022tgl} & CTDG & single-node multi-GPU & DGL \\
DistTGL~\cite{zhou2023disttgl} & CTDG & multi-node multi-GPU & DGL \\
SPEED~\cite{chen2023speed} & CTDG & single-node multi-GPU & Pytorch \\
Xia et al.~\cite{xia2023redundancy} & CTDG & single-node multi-GPU & PyTorch \\
STEP~\cite{li2023step} & CTDG & single-node single-GPU with multi-core CPU & PyTorch \\ \hline
\end{tabular}
}
\end{table*}

To further enhance efficiency, DistTGL~\cite{zhou2023disttgl} implements a mechanism where memory operations are serialized and executed asynchronously through a separate daemon process, circumventing intricate synchronization requirements. DistTGL~\cite{zhou2023disttgl} incorporates an additional static node memory within the dynamic GNNs, which not only enhances the model's accuracy but also accelerates its convergence rate. Two novel training methodologies, namely time period parallelism and memory parallelism, are also proposed. These two methodologies enable capturing a comparable number of graph event dependencies on multiple GPUs, similar to that achieved in single GPU training. Prefetching and pipelining techniques are utilized to mitigate the associated costs of batch formation and to facilitate overlap with GPU training. The DistTGL~\cite{zhou2023disttgl} framework offers notable enhancements in terms of convergence speed and training throughput compared to TGL~\cite{zhou2022tgl}. This enables the efficient scaling of memory-based dynamic GNNs training on distributed GPU clusters, achieving near-linear speedup ratios. Guidance on the identification of the most effective training configurations is provided by employing a variety of parallelization algorithms that are tailored to the specific dataset and hardware attributes.

In contrast, the SPEED~\cite{chen2023speed} framework introduces a streaming edge partitioning module that integrates temporal information through an exponential decay technique. Additionally, it effectively minimizes the replication ratio by regulating the number of shared nodes. A vertex partitioning method based on interval properties is also employed, combined with joint and asynchronous pipelining techniques between mini-batches. This ensures efficient parallel training on multiple GPUs for large-scale dynamic graphs. The design speeds up the training and reduces memory consumption on a single GPU significantly.
% Demirci et al.~\cite{demirci2022scalablegcn} introduce a sparse matrix partitioning scheme based on the hypergraph partitioning model. By measuring communication costs, they can reduce the overall communication overhead effectively. The fundamental principle of this method lies in the use of peer-to-peer communication, in contrast to aggregate communication, ensuring only the most vital data is transmitted. Additionally, the parameter matrix is replicated to each processor to improve computational locality. This approach outperforms traditional single-node GCN implementations on large-scale distributed-memory systems, especially when dealing with large graphs that have a relatively low average node degree.

Liu et al.~\cite{xia2023redundancy} proposed a method that emphasizes hierarchical pipeline parallelism, integrating data prefetching, joint pipelining across mini-batches, and asynchronous pipelining. Additionally, to enhance communication efficiency, they introduced a strategy for partitioning graphs and vertices to eliminate redundancy. 
Another framework, STEP~\cite{li2023step}, provides an unsupervised pruning method for large-scale dynamic graphs. Node and edge representations are acquired by self-supervised adversarial learning subsequent to each graph update. Moreover, it selects edges for sampling depending on their significance, resulting in a graph known as the undermine graph. STEP utilizes the graph provided to train a basic pruning network. This enables the STEP framework to effectively manage the process of removing unnecessary elements from dynamic graphs, during both the training and inference stages, with a particular focus on newly added edges.

In general, these advanced methods jointly enhance the effective parallel training of dynamic graph neural networks in distributed settings, each with its distinct benefits and potential use cases, as shown in Table~\ref{tab:training_frameworks}.
}

\section{Datasets and Benchmarks Overview}
\label{sec:dataset}
% \subsection{Common Tasks and Commonly Used Evaluation Metrics}
% \textbf{Dynamic Node Classification} \\
% \textbf{Temporal Link Prediction} \\
% \textbf{Time Prediction}
\subsection{Datasets}
\begin{table*}[!t]
\centering
\caption{The statistics of datasets, where $n$ represents the number of nodes, $m$ denotes the number of edges, $|T|$ indicates the number of timestamps or snapshots, and $d_e$ refers to the dimension of edge features.}
\label{tab:dataset}
\begin{threeparttable}
\begin{tabular}{l l ll l l l l}
\hline
Types & Dataset &  Domain 
&Category & $n$ & $m$ & $|T|$ & $d_e$ \\ \hline
\multirow{18}{*}{CTDG} 
 & Social Evolution~\cite{data_socialwvolution} &  Proximity &general & 74 & 2,099,519 & 565,932 & 1\textsuperscript{†}  \\
 & Enron~\cite{data_enron} &  Social &general & 184 & 125,235 & 22,632 &  -\\
 & Contact~\cite{data_contact} &  Proximity &general & 692 & 2,426,279 & 8,065 & 1\textsuperscript{†}\\
 & UCI~\cite{data_uci} &  Social &general & 1,899 & 59,835 & 58,911 & 100 \\
 & LastFM~\cite{kumar2019jodie} &  Interaction &bipartite & 1,980 & 1,293,103 & 1,283,614 &  -\\
 & Bitcoin-Alpha~\cite{data_bitcoins1, data_bitcoins2} &   Finance &general& 3,783 & 24,186 & 24,186 &  -\\
 & Bitcoin-OTC~\cite{data_bitcoins1, data_bitcoins2} &   Finance &general& 5,881 & 35,592 & 35,592 &  -\\
 & MOOC~\cite{kumar2019jodie} &  Interaction &bipartite & 7,144 & 411,749 & 345,600 & 4\textsuperscript{‡}\\
 & Wikipedia~\cite{kumar2019jodie} &  Social &bipartite & 9,227 & 157,474 & 152,757 & 172\\
 & Reddit~\cite{kumar2019jodie} &  Social &bipartite & 10,984 & 672,447 & 669,065 & 172\\
 & GDELT~\cite{zhou2022tgl, data_gdelt} &  Event &general & 16,682 & 191,290,882 & 170,522 & 186 \\ 
 & eBay-Small~\cite{huang2023benchtemp} &  E-commerce &bipartite & 38,427 & 384,677 & - &  -\\
 & Taobao~\cite{data_taobao1, data_taobao2} &  E-commerce &bipartite & 82,566 & 77,436 & - & 4\textsuperscript{‡}\\
 & YouTube-Reddit-Small~\cite{data_youtubereddit} &  Social &bipartite & 264,443 & 297,732 & - &  -\\
 & eBay-Large~\cite{huang2023benchtemp} &  E-commerce &bipartite & 1,333,594 & 1,119,454 & - &  -\\
 & Taobao-Large~\cite{data_taobao1, data_taobao2} &  E-commerce &bipartite & 1,630,453 & 5,008,745 & - & 4\textsuperscript{‡}\\
 & DGraphFin~\cite{data_dgraphfin} &  Social &general & 3,700,550 & 4,300,999 & - &  -\\
 & Youtube-Reddit-Large~\cite{data_youtubereddit} &  Social &bipartite & 5,724,111 & 4,228,523 & - &  -\\
 \hline
\multirow{10}{*}{DTDG} 
 & UN Vote~\cite{data_unvote} &  Politics &general & 201 & 1,035,742 & 72 & 1\textsuperscript{†}\\
 & US Legislative~\cite{data_canparl, data_uslegislative} &  Politics &general & 225 & 60,396 & 12 & 1\textsuperscript{†}\\
 & UN Trade~\cite{data_untrade} &  Finance &general & 255 & 507,497 & 32 & 1\textsuperscript{†}\\
 & Canadian Parliament~\cite{data_canparl} &  Politics &general & 734 & 74,478 & 14 &  -\\
 & Twitter-Tennis~\cite{data_twittertennis} &  Social & general& 1,000 & 40,839 & 120 &  -\\
 & Autonomous systems~\cite{data_as733} &   Communication &general & 7,716 & 13,895 & 733 &  - \\
 & Flights~\cite{data_flights} &  Transport &general & 13,169 & 1,927,145 & 122 & 1\textsuperscript{†}\\
 & HEP-TH~\cite{data_as733, data_hepth} & Citation & general & 27,770 & 352,807 & 3,487 &  -\\
 & Elliptic~\cite{data_elliptic} &   Finance &general & 203,769 & 234,355 & 49 &  - \\
 & MAG~\cite{zhou2022tgl, hu2021ogb} &  Citation &general & 121,751,665 & 1,297,748,926 & 120 &  -\\
\hline
\end{tabular}
\begin{tablenotes}
    \item[†] Represents the edge weight.
    \item[‡] Indicates the number of different behaviors.
\end{tablenotes}
\end{threeparttable}
\vspace{-2mm}
\end{table*}
In recent years, the quantity and variety of dynamic graph datasets made available to the public have increased significantly. These datasets provide valuable resources for researchers to validate and compare algorithms, thereby facilitating the exploration of new research directions. Table~\ref{tab:dataset} contains basic characteristics of commonly used dynamic graph datasets. We can divide these datasets into CTDGs and DTDGs, according to the collection and storage methods of the datasets. However, with appropriate processing, they can be transformed into one another. From these datasets, we can observe the following characteristics:
\begin{itemize}
\item \textbf{Domains and applications.} These datasets cover a diverse range of domains, including social, music, political, transportation, economic, and e-commerce networks. This diversity enables researchers to test and validate algorithms on various real-world applications.
\item \textbf{Scale.} The datasets vary in graph sizes, ranging from small-scale graphs with thousands of nodes and edges (e.g., UCI~\cite{data_uci} and Contact~\cite{data_contact}) to large-scale graphs with millions or even billions of nodes (e.g. Taobao~\cite{data_taobao1, data_taobao2} and MAG~\cite{zhou2022tgl}). This indicates that existing datasets can support research and applications at different scales.
\item \textbf{Types.} Many datasets are bipartite graphs. For example, datasets such as MOOC~\cite{kumar2019jodie}, LastFM~\cite{kumar2019jodie}, eBay~\cite{huang2023benchtemp}, and TaoBao~\cite{huang2023benchtemp} fall into this category, with nodes categorized into two distinct sets and edges representing interactions between these sets. In contrast, some datasets, like Enron~\cite{data_enron} and SocialEvo~\cite{poursafaei2022edgebank}, are general graphs.
\item \textbf{Features.} Although these datasets provide rich information about nodes and edges, many of them do not provide features of nodes or edges. Only a few datasets like Reddit~\cite{kumar2019jodie}, Wikipedia~\cite{kumar2019jodie} and UCI~\cite{data_uci} provide edge features, which can help researchers gain deeper insights into interaction patterns between nodes. Datasets offering node features are even rarer, with GDELT~\cite{zhou2022tgl}, Elliptic~\cite{data_elliptic}, and MAG~\cite{zhou2022tgl} being notable exceptions. Conversely, datasets such as LastFM~\cite{kumar2019jodie}, Enron~\cite{data_enron}, and Flights~\cite{data_flights} lack node feature details, potentially due to privacy concerns or the sensitive nature of the raw data. Moreover, in practical applications, identifying or defining appropriate features can also be challenging.
\end{itemize}
Overall, the currently available public datasets provide diverse resources for representation learning on dynamic graphs. However, these datasets also exhibit some common characteristics, such as the lack of node and edge features, which pose challenges and opportunities for researchers. This is because collecting features in real-world applications is often costly, and learning patterns solely from graph structure changes can improve model generalization. It would be very encouraging if new high-quality public datasets with extensive features were to emerge in the future, as this would facilitate more research projects.

\subsection{Prediction Problems}
Graph learning tasks include three levels: node-level, edge-level, and graph-level~\cite{hu2021ogb}. In node-level tasks, the primary focus is on the attributes of individual nodes, which include node classification, regression, clustering, and so on. Edge-level tasks emphasize information about the edge. For example, link prediction determines whether an edge exists between two nodes, while link classification identifies the type or attributes of an edge. Graph-level tasks analyze the entire structure, such as classifying the category of the graph (graph classification) or predicting its continuous values (graph regression). Graph-generation tasks aim to generate new graphs. In addition, there are specialized tasks, such as graph generation, whose objective is to generate entirely new graph structures based on specific criteria or patterns observed in existing data.

The three levels of tasks also exist in dynamic graphs. The primary difference from the static context is that the graph changes over time, resulting in simultaneous alterations to the properties of nodes, edges, and the entire graph. For node-level tasks, attributes and connection patterns of nodes may change, potentially impacting the results of classification or regression analyses. In edge-level tasks, the presence and attributes of edges may shift over time, for example, new connections can emerge while old ones disappear. For graph-level tasks, the topology of the graph might become more intricate with the addition of new nodes and edges, which then impact results in graph classification, regression, and generation. Next, we will explore specific application scenarios of dynamic graph analytics using link prediction, anomaly detection, and community discovery as examples.

\noindent \textbf{Link prediction} in dynamic graphs aims to predict potential edges in future graphs based on connections from previous time steps, also known as temporal link prediction. This requires not only the structural features of the current graph but also identifying trends from its dynamic changes. Link prediction has a wide range of applications. In citation networks, by examining both authors' evolving citation behaviors and changes in research focus, we can predict the authors they might reference in future publications.

\noindent \textbf{Anomaly detection} in dynamic graphs can be categorized into edge and node anomaly detection. Edge anomaly detection aims to identify edges in a dynamic graph that deviate from historical evolution patterns. Such irregularities may arise from connections that rogue users have introduced or manipulated maliciously. Conversely, node anomaly detection focuses on identifying nodes that distinctly differ from their peers or exhibit atypical behaviors. Anomaly detection has broad applications, particularly in fraud detection and e-commerce. For example, on e-commerce sites, users might manipulate the system by repeatedly clicking on specific products to increase their popularity artificially. They could also simultaneously click on a product and another trendy item, deceiving the recommendation system into overestimating the similarity between the two items. These connections are considered anomalous edges, and the deceptive users are considered anomalous nodes.

\noindent \textbf{Community detection} is a fundamental task in understanding complex graphs. Community structures are commonly observed in real-world networks that encompass various domains, such as social, biological, or technological. Based on the density of connections among nodes, the graph can be divided into various communities. Typically, nodes within a community have denser connections, while connections between nodes from different communities are less frequent. This community structure is also present in dynamic graphs, but the community configurations can evolve as the graph structure changes. The identification and comprehension of dynamic community structures and their evolutionary patterns can provide a deeper comprehension of the graph and its community formations, which facilitates the prediction of potential changes in the graph. Taking social networks as an example, users with similar likes and follows can be grouped into the same interest clusters. By identifying these clusters, the system is able to recommend new connections or content that aligns with their interests.
\begin{table*}[t]
\centering
\caption{Links of benchmarks}
\label{tab:benchmarks}
{\tiny
\begin{tabular}{l l}  \hline
Benchmark & Code \\ \hline
PyTorch Geometric Temporal & \href{https://github.com/benedekrozemberczki/pytorch_geometric_temporal}{https://github.com/benedekrozemberczki/pytorch\_geometric\_temporal} \\
TGL & \href{https://github.com/amazon-science/tgl}{https://github.com/amazon-science/tgl} \\
Dynamic Graph Library (DyGLib) & \href{https://github.com/yule-BUAA/DyGLib}{https://github.com/yule-BUAA/DyGLib} \\
Temporal Graph Benchmark (TGB) & \href{https://github.com/shenyangHuang/TGB}{https://github.com/shenyangHuang/TGB } \\
BenchTemp & \href{https://github.com/qianghuangwhu/benchtemp}{https://github.com/qianghuangwhu/benchtemp} \\
\hline
\end{tabular}
}
\end{table*}

\noindent \textbf{Transductive / inductive setting.} In the transductive setting, the model attempts to make accurate predictions on the graph on which it was initially trained. It trains on a specific portion of the graph while being fully aware of the entire structure, including the unlabeled portions. In contrast, in the inductive setting, the model learns from a designated training graph to make reliable predictions on independent, completely unknown graphs. In dynamic graphs, where nodes and edges frequently change over time, the essence of the inductive setting is the model’s ability to effectively generalize to unseen nodes and dynamically evolving structures.

\noindent \textbf{Interpolation / extrapolation.} When predicting the value of a function in an unknown region based on existing data and models, predictions within the range of existing data are called interpolations, while those outside are termed extrapolations. Taking dynamic node classification as an example, extrapolation involves predicting future states $(>t)$ based on observations up to time $t$, such as in weather forecasting. The interpolation setting involves predicting states within the observed range $(t_0, t)$ based on observations up to time $t$, like completing missing values. Similarly, in link prediction, extrapolation predicts whether future edges $(v_i, v_j)$ will exist, while interpolation determines if edges $(v_i, v_j)$ exist within the observed range $(t_0, t)$, as in knowledge graph completion. In summary, interpolation involves predicting unobserved values within existing data, whereas extrapolation predicts beyond the observed range. For dynamic graphs, extrapolation is used for future predictions, while interpolation focuses on imputing missing information within observations, depending on whether the target concerns observed or unobserved future states.

\subsection{Benchmarks}
In the past, different methods had their own configurations and implementations, complicating the direct comparison of model performances. Recent work has tried to establish unified frameworks for training and evaluation to rectify this. By providing standardized datasets, synchronized preprocessing stages, and fair training strategies, these frameworks make sure that all methods are tested in the same way. Researchers can then easily compare different models for the same tasks, pinpointing the merits and shortcomings of each approach with greater clarity. These efforts demonstrate a prevailing tendency towards heightened standardization in the field of dynamic graphs, which benefits the evaluation and comparison of different methods. Table~\ref{tab:benchmarks} provides links to these benchmarks.

% \section{Optimization Techniques and Practices}
% \subsection{Applications of Transfer Learning and Pre-trained Models in Dynamic Graphs}
% \textbf{Dynamic Transfer Learning.} 
\section{\revision Advanced Techniques in Graph Processing}
\label{sec:new_techniques}
\subsection{Dynamic Transfer Learning}
Transfer learning employs the knowledge of a model trained on one task (called the source task) to facilitate the learning of another distinct but related task (called the target task). The core idea is to share specific knowledge between the source and the target task. This concept is based on the observation that various tasks frequently share underlying features or patterns. Consequently, insights from one task can potentially benefit another~\cite{pan2009transfer, neyshabur2020transfer}. Dynamic transfer learning is a transfer learning setting proposed for scenarios where both the source and target domains evolve. In this context, the current state of the target domain depends on its historical state, which cannot be transferred straightforwardly. Thus, dynamic transfer learning requires continuous real-time adjustments to the shifting data distributions in both the source and target domains. It also demands the prompt transfer of pertinent knowledge from the evolving source domain to the concurrently changing target domain, aiming to optimize performance in the target tasks. The primary challenge lies in effectively modeling inter-domain evolution and managing temporal interdependencies across various time steps.

Dynamic transfer learning was first introduced to address diverse real-world challenges in the areas of computer vision and natural language processing. Recently, researchers have tried to explore its potential in graph learning. As of now, research into dynamic transfer learning for graphs is still in its early stages. It mainly involves applying ideas from fields other than graphs, like using LSTM or attention mechanisms to model the graph's temporal pattern~\cite{wang2023dynamictrans}. In domains outside of graph theory, substantial research has been conducted on dynamic source domain non-stationarity, target domain concept drift, and related topics. Future investigations should delve deeper into the specific attributes of graph data, aiming to design more suitable dynamic transfer learning methods. This will further position dynamic transfer learning as a pivotal tool in addressing various time-sensitive practical challenges.

\subsection{Pretraining Techniques for Dynamic GNNs}
% \noindent \textbf{Pre-Training on Dynamic GNNs.} 
Typically, pre-trained models are trained on large-scale datasets to obtain universal feature representations, which are then fine-tuned on specific downstream tasks to enhance their performance. Originating from the fields of computer vision and natural language processing, pre-trained models have achieved significant successes in many tasks, such as semantic parsing, image classification and image segmentation. With the increase in the volume of graph data, pre-training models have recently been introduced to the domain of graph learning. Related work usually employs self-supervised strategies to design pre-training tasks such as edge generation~\cite{hu2020gptgnn}, attribute masking~\cite{hu2020gptgnn} and subgraph comparison~\cite{qiu2020gcc}. Since nodes and edges in dynamic graphs vary over time, these models face the issue of describing the temporal dependence. Therefore, PT-DGNN~\cite{chen2022ptdgnn} uses a time-based masking strategy to predict the most recent edges or attributes, learning the temporal evolution law of the graph. On the other hand, CPDG~\cite{bei2023cpdg} contrasts pairs of positive and negative samples. CPDG captures patterns both temporally and structurally by enhancing the similarity between positive samples and diminishing it among negative samples. Additionally, CPDG stores memory states from different time steps to help the model preserve long-term evolutionary information, further supporting subsequent downstream tasks.

However, there are still numerous challenges that require additional research. Scalability is a major issue in this field, and the key question is how to train models in advance so that they can efficiently deal with dynamic graphs at different temporal granularities and still be flexible enough to perform many different downstream tasks. In addition, dynamic graphs contain both long-term stability patterns and short-term variability patterns, which means the model needs to adjust for both. Furthermore, the structural neighbor information of each node is essential, so the model is required to preserve this structural information effectively. In real-world applications, pre-training for large-scale dynamic graphs requires efficient sampling and training mechanisms. Simultaneously, the temporal patterns learned by the pre-trained models should be promptly incorporated into subsequent tasks.

In summary, current research and applications primarily focus on designing appropriate pre-training tasks, sampling strategies, and contrastive learning methods for dynamic graphs. The objective is to achieve resilient temporal and structural depictions that improve the adaptability of the model for various tasks. Simultaneously, it is important to include enduring and immediate trends effectively, and then improve the efficacy of model training and transfer.
\section{Future Trends and Challenges}
\label{sec:challenge}
\subsection{More Complex Dynamic Graphs}
Although some large-scale dynamic graph datasets have emerged in recent years, dynamic graph datasets are still lacking in diversity compared to static graph datasets. This may lead to insufficient model performance in diverse, real-world, dynamic graph applications.
\begin{itemize}
\item Firstly, the results of EdgeBank~\cite{poursafaei2022edgebank} indicate that in some cases, using only historical information can achieve good results. This suggests that the complexity of current datasets may be insufficient to distinguish between different models significantly. Therefore, we need to construct more challenging datasets containing richer structural information and temporal evolution patterns to evaluate models' modeling capabilities better.
\item Secondly, operations like users unfollowing others are shared in real applications, but currently, there is a lack of real-world datasets containing such edge deletion data. Expanding datasets could involve collecting and integrating data with real edge deletion operations to simulate the dynamics of graph data in the real world. This can help assess models' robustness and performance when handling edge deletion.
\item Finally, some special graphs are rarely mentioned in current works, such as dynamic signed networks~\cite{sharma2022signed} and dynamic heterophilic graphs. In the future, dynamic graph datasets containing different network topologies (e.g. small-world, scale-free) could be constructed, and more special graphs could be collected and generated.
\end{itemize}
In summary, larger-scale, richer-content, and more complex structure datasets are needed to facilitate the development of dynamic graph learning towards handling real-world complex networks.

% \subsection{Multimodal and Cross-domain Learning}
\subsection{Interpretability of Dynamic GNNs}
% \subsection{Trend Prediction}
The frameworks and methods related to deep model interpretability are designed to help users understand and trust these models, enabling them to be applied more effectively. There have been some advances in the area of interpretability for static GNN models~\cite{dai2021selfexplainable}, while research on interpretability for dynamic GNNs appears to still be in its early stages. Xie et al.~\cite{xie2022dgexplainer} proposed DGExplainer, which aims to provide a reliable interpretation of dynamic GNNs. DGExplainer redistributes the output activation scores of a dynamic GNN to the neurons of the previous layer to calculate the relevance scores of the input neurons, thereby identifying important nodes in link prediction and node regression tasks. The DynBrainGNN~\cite{zheng2023dynbraingnn} model, designed to provide both prediction and explanation, contains built-in interpretable GNNs with dynamic FCs for psychiatric disease analysis. These studies show that although the interpretability of dynamic GNNs is a relatively new and less discussed area, there are already some studies attempting to address this issue.

{\revision
\subsection{LLMs for dynamic graph learning}
\label{sec:LLMs_for_dynamicgraphs}
Recently, large language models (LLMs) have shown remarkable capabilities in natural language processing, achieving breakthroughs in many aspects, such as text generation~\cite{NEURIPS2020_chatgpt}, question answering~\cite{zhuang2024llmqa}, and dialogue tasks~\cite{mendoncca2023llmdialogue}. Researchers have tried to adapt LLM to more fields to utilize its powerful pre-trained language understanding ability. LLM has strong few-shot, in-context learning ability, and by designing prompting techniques carefully, it can help dynamic graph learning models to understand spatial and temporal information better. For example, LLM4DyG~\cite{zhang2023llm4dyg} examines that providing only one example can significantly improve the accuracy of LLM on multiple dynamic graph tasks such as “when link", “when connect", “when triadic closure", “neighbor at time" and “neighbor in periods". On the other hand, LLM can perform reasoning in a given context and solve dynamic graph related tasks. This insight motivates new applications of LLMs in dynamic graph learning tasks and potentially transforms approaches in related fields. Although LLMs have demonstrated inspiring understanding capabilities on dynamic graph tasks, they still face challenges as follows.
\begin{itemize}
    \item \textbf{Processing large-scale graphs}: As graph size grows, the performance of LLMs markedly declines~\cite{tang2023graphgpt}. The reason may be that the increase in graph size makes the amount of contextual information soaring, which makes it difficult for LLM to capture the key information. The ability of LLMs to model large-scale dynamic graphs remains to be validated.
    \item \textbf{Computational efficiency}: Current LLMs are mostly large and computationally expensive. How to improve the computational speed of LLMs and improve their efficiency in learning dynamic graphs is an issue that needs to be addressed.
    \item \textbf{Lack of structural modeling}: LLMs lack the capability to explicitly model graph structures in the same way that GNN does. Merely relying on contextual descriptions to acquire information may not be adequate and direct. Therefore, additional investigation is required to improve LLMs' ability to model graph topology.
\end{itemize}

}

\section{Conclusion}
\label{sec:conclusion}
% \subsection{Main Findings and Summary}
% \subsection{Recommendations for Future Research}
Dynamic graph neural networks, a recent development in representation learning, can process \eat{diverse real-time }changing graph structures and have potential applications across various fields. In this paper, we conduct a comprehensive survey on the key issues of dynamic GNNs, including prediction tasks, temporal modeling methods, and the handling of large-scale dynamic graphs. We have summarized the main advancements in current research and pointed out some promising key directions, such as expanding the scope of existing models and researching the expressive capability of dynamic GNNs. Dynamic GNNs still face challenges, such as scalability, theoretical guidance, and dataset construction. We hope that this paper can provide valuable references for researchers in this field, and promote the further development of theories and methods related to dynamic GNNs. An important direction for future research is to continue expanding the applicability of dynamic graph representation learning, handling more complex and diverse dynamic graph structures and evolution patterns, and making it a unified representation learning framework with both breadth and depth. As more and more practical applications involve dynamic graph analysis, dynamic GNNs will undoubtedly become a continuously active and increasingly important research area.

\begin{acknowledgement}
% The ``Acknowledgements'' section is the general term for the list of contributions, credits, and other information included at the end of the text of a manuscript but before the references. Authors should obtain written permission to include the names of individuals in the ``Acknowledgements'' section.
% \todo{Check.} 
This research was supported in part by National Science and Technology Major Project (2022ZD0114 802), by National Natural Science Foundation of China (No. U2241212, No. 61932001), by Beijing Natural Science Foundation (No. 4222028), by Beijing Outstanding Young Scientist Program No.BJJWZYJH012019100020098, by Alibaba Group through Alibaba Innovative Research Program, and by Huawei-Renmin University joint program on Information Retrieval. We also wish to acknowledge the support provided by the fund for building world-class universities (disciplines) of Renmin University of China, by Engineering Research Center of Next-Generation Intelligent Search and Recommendation, Ministry of Education, Intelligent Social Governance Interdisciplinary Platform, Major Innovation \& Planning Interdisciplinary Platform for the “Double-First Class” Initiative, Public Policy and Decision-making Research Lab, and Public Computing Cloud, Renmin University of China. The work was partially done at Beijing Key Laboratory of Big Data Management and Analysis Methods, MOE Key Lab of Data Engineering and Knowledge Engineering, and Pazhou Laboratory (Huangpu), Guangzhou, Guangdong 510555, China.
\end{acknowledgement}

\bibliographystyle{fcs}
\bibliography{ref}

% \vspace{-6mm}
\begin{biography}{photo_zyp}
Yanping Zheng is a Ph.D. candidate at Gaoling School of Artificial Intelligence, Renmin University of China, advised by Professor Zhewei Wei. She received her master's degree of engineering from Beijing Technology and Business University in 2020. Her research focuses on graph learning algorithms. She is particularly interested in efficient algorithms on Graph Neural Networks, Dynamic Graph Representation Learning.
\end{biography}

% \vspace{-2mm}
\begin{biography}{photo_yl}
Lu Yi is currently a Ph.D. student at Gaoling School of Artificial Intelligence, Renmin University of China, advised by Professor Zhewei Wei. She received her B.E. degree in Computer Science and Technology at School of Computer Science, Beijing University of Posts and Telecommunications in June 2022. Her research lie in the field of graph-related machine learning and efficient graph algorithm.
\end{biography}

% \vspace{-2mm}
\begin{biography}{photo_wzw}
Zhewei Wei is currently a Professor at Gaoling School of Artificial Intelligence, Renmin University of China. He obtained his Ph.D. degree at Department of Computer Science and Engineering, HKUST in 2012. He received the B.Sc. degree in the School of Mathematical Sciences at Peking University in 2008. His research interests include graph algorithms, massive data algorithms, and streaming algorithms. He was the Proceeding Chair of SIGMOD/PODS2020 and ICDT2021, the Area Chair of ICML 2022/2023, NeurIPS 2022/2023, ICLR 2023, WWW 2023. He is also the PC member of various top conferences, such as VLDB, KDD, ICDE, ICML and NeurIPS.
\end{biography}

% \clearpage
% \appendix
% \setcounter{table}{0}
% \setcounter{figure}{0}
% \renewcommand{\thetable}{A.\arabic{table}}
% \renewcommand{\thefigure}{A.\arabic{figure}}
% \input{author_response}

\end{document}